\documentclass{article}
\usepackage{arxiv}

\usepackage[utf8]{inputenc} 
\usepackage[T1]{fontenc}    
\usepackage{hyperref}       
\usepackage{url}            
\usepackage{booktabs}       

\usepackage{times}  
\usepackage{helvet}  
\usepackage{courier}  
\usepackage{graphicx} 
\usepackage{natbib}  
\usepackage{caption} 
\DeclareCaptionStyle{ruled}{labelfont=normalfont,labelsep=colon,strut=off} 
\usepackage{algorithm}
\usepackage{algorithmic}
\relax

\usepackage{amsmath,amssymb}
\usepackage{microtype}

\usepackage[usenames,dvipsnames]{color}

\usepackage{todonotes}

\definecolor{darkgreen}{rgb}{0,0.5,0}

\usepackage{siunitx}
\DeclareMathOperator{\GammaDist}{Gamma}

\usepackage{newfloat}
\usepackage{listings}
\floatstyle{ruled}
\newfloat{listing}{tb}{lst}{}
\floatname{listing}{Listing}

\usepackage[utf8]{inputenc} 
\usepackage[T1]{fontenc}    
\usepackage{booktabs}       
\usepackage{amsfonts}       
\usepackage{nicefrac}       
\usepackage{microtype}      
\usepackage{soul}           
\usepackage{amssymb}       

\usepackage{amsmath}

\usepackage{array}
\usepackage{multirow}
\usepackage{multicol}
\usepackage{booktabs}
\usepackage{amsmath}
\newcolumntype{M}[1]{>{\centering\arraybackslash}m{#1}}
\usepackage{graphicx}
\usepackage{graphics}

\usepackage{comment}
\usepackage{soul} 
\usepackage{listings}
\usepackage{float}
\usepackage{placeins}
\usepackage{subcaption}
\usepackage{wrapfig,lipsum,booktabs}
\usepackage{natbib}
\usepackage[normalem]{ulem}
\usepackage{bbm}
\usepackage{amsthm}
\usepackage{amsmath}
\usepackage{appendix}
\usepackage{cleveref}

\usepackage[algo2e]{algorithm2e}
\usepackage{algorithm}
\usepackage{accents}

\usepackage{thmtools,thm-restate}
\usepackage{enumitem}

\usepackage{xcolor}

\newcommand{\bi}{\begin{itemize}}
\newcommand{\ei}{\end{itemize}}
\newcommand{\BE}{\begin{enumerate}}
\newcommand{\EE}{\end{enumerate}}

\newcommand{\eg}{\mbox{\it e.g.}}

\renewcommand\cite{\citep}	

\newcommand{\assignmat}{\mathbf{X}}

\newcommand{\assignmatentry}{x}

\newcommand{\papers}{\mathcal{P}}
\newcommand{\reviewers}{\mathcal{R}}
\newcommand{\pcs}{\mathcal{PC}}

\newcommand{\spcs}{\mathcal{SPC}}
\newcommand{\acs}{\mathcal{AC}}
\newcommand{\obj}{O}

\newcommand{\scorematrixentry}{s}

\newcommand{\subjectareas}{\Psi}
\newcommand{\subjectarea}{\psi}

\newcommand{\capacity}{c}
\newcommand{\capacityentry}{c}
\newcommand{\lowerlimitcapacity}{l}
\newcommand{\capacityslack}{s^{\text{cap}}}

\newcommand{\capacityslackpenalty}{p^{\text{cap}}}

\newcommand{\cyclevars}{s^{\text{cy}}}
\newcommand{\cyclepen}{p^{\text{cy}}}

\newcommand{\cycles}{CY}

\newcommand{\seniorityslack}{s^{\text{sen}}}
\newcommand{\seniorityreward}{\text{Reward}^{\text{sen}}}

\newcommand{\numregion}{\text{reg}}
\newcommand{\paperregion}{v}
\newcommand{\region}{r}
\newcommand{\regions}{\text{Regions}}
\newcommand{\reviewerregion}{t}
\newcommand{\regionreward}{\text{Reward}^\text{reg}}

\newcommand{\coauthordist}{d}

\newcommand{\coreview}{\text{cad}}
\newcommand{\coreviewerset}{V}

\newcommand{\coimatentry}{\text{coi}}

\newcommand{\constraints}{\mathcal{C}}

\title{Matching Papers and Reviewers at Large Conferences}
\author{
  Kevin Leyton-Brown \\
  Department of Computer Science\\
  University of British Columbia\\
  Vancouver, Canada\\
  \texttt{kevinlb@cs.ubc.ca} \\
  \And
  Mausam \\
  Department of Computer Science and Engineering \\
  Indian Institute of Technology Delhi \\
  New Delhi, India \\
  \texttt{mausam@cse.iitd.ac.in} \\
  \And
  Yatin Nandwani\thanks{Joint student lead authors.}\\
  Department of Computer Science and Engineering \\
  Indian Institute of Technology Delhi \\
  New Delhi, India \\
  \texttt{yatin.nandwani@cse.iitd.ac.in} \\
  \And
  Hedayat Zarkoob\footnotemark[1] \\
  Department of Computer Science\\
  University of British Columbia\\
  Vancouver, Canada\\
  \texttt{hzarkoob@cs.ubc.ca} \\
  \And
  Chris Cameron \\
  Department of Computer Science\\
  University of British Columbia\\
  Vancouver, Canada\\
  \texttt{cchris13@cs.ubc.ca} \\
  \And
  Neil Newman \\
  Department of Computer Science\\
  University of British Columbia\\
  Vancouver, Canada\\
  \texttt{newmanne@cs.ubc.ca } \\
  \And
  Dinesh Raghu \\
  IBM Research \\
  New Delhi, India \\
  \texttt{diraghu1@in.ibm.com} \\
}

\date{\vspace{-1em}}
\begin{document}


\maketitle

\begin{abstract}

Peer-reviewed conferences, the main publication venues in CS, rely critically on matching highly qualified reviewers for each paper. Because of the growing scale of these conferences, the tight timelines on which they operate, and a recent surge in explicitly dishonest behavior, there is now no alternative to performing this matching in an automated way. 
This paper studies a novel reviewer--paper matching approach that was recently deployed in the 35th AAAI Conference on Artificial Intelligence (AAAI 2021), and has since been adopted (wholly or partially) by other conferences including ICML 2022, AAAI 2022, and IJCAI 2022. This approach has three main elements: (1) collecting and processing input data to identify problematic matches and generate reviewer--paper scores; (2) formulating and solving an optimization problem to find good reviewer--paper matchings; and (3) a two-phase reviewing process that shifts reviewing resources away from papers likely to be rejected and towards papers closer to the decision boundary.
This paper also describes an evaluation of these innovations based on an extensive post-hoc analysis on real data---including a comparison with the matching algorithm used in AAAI's previous (2020) iteration---and supplements this with additional numerical experimentation.\footnote{Writing this paper and conducting the experimental analysis described herein took almost a full year beyond the conclusion of AAAI 2021. Nevertheless, this paper is grounded in our experiences as Program Chairs (Mausam, Kevin), Workflow Chairs (Hedayat, Dinesh) and Associate Workflow Chairs (Yatin, Chris, Neil) at AAAI 2021. Many other people contributed to the the conference's design and operation. Notably, Qiang Yang was Conference Chair; Gabriele Röger and Yan Liu were Associate Program Chairs, and Guangneng Hu was a Workflow Chair; all contributed to the design and planning process throughout the year preceding the conference. 
Even more concretely impacting the content of this paper, Gabi provided draft text on filtering manipulative bids that we adapted into Section \ref{sec:filter-manipulative-bids}, and Yan made contributions to the MIP formulation described in Section \ref{sec: MIP}. Keshav Sai Kolluru helped in the analysis of the bids, and Shantanu Agarwal helped in coding Subject Area Matching Score (\cref{subsec:SAMS}).
The CMT support team provided invaluable real-time service and development during the conference which made our introduction of two-phase reviewing process feasible. Carol Hamilton and the AAAI Office provided innumerable forms of behind-the-scenes support. The AAAI 2021 conference and executive committees supported us in experimenting with our unconventional ideas for the conference. Finally, an organization as large as AAAI operates only through the efforts of literally thousands of volunteers in myriad roles; we thank all of them  for their continued service to the community. }
\end{abstract}



\section{Introduction}
\label{sec: intro}
\setcounter{footnote}{1}


Submissions to prominent AI conferences have grown steadily in recent years. The AAAI Conference on Artificial Intelligence (AAAI) received fewer than 1,000 submissions in each of its first 33 years (1980--2012); fewer than 2,000 submissions annually until 2015; and fewer than 4,000 submissions annually until 2018. AAAI 2021 received over 9,000 submissions! Program committees have grown to keep pace with submissions; nearly 10,000 reviewers were involved in the 2021 conference. Given the scale and the tight timeline on which such large conferences operate---roughly three months between paper submission and author notification---it is becoming an increasingly challenging problem to assign reviewers\footnote{Modern conference reviewing is often divided into four hierarchical roles, though terminology can vary. At AAAI, Program Committee members (PCs) review submissions; Senior Program Committee members (SPCs) facilitate discussions among PCs and write meta-reviews; Area Chairs (ACs) oversee the review process, communicate with SPCs about final decisions, and recommend accept/reject decisions; Program Chairs design and oversee the whole review process and make final decisions for each submission. In this paper, we sometimes use the term reviewer to refer exclusively to PCs; in other cases, we refer to PCs, SPCs, and ACs. Distinctions between the two cases should nevertheless be clear from the context.} to papers about which they can provide high-quality reviews. Some key subproblems are estimating how qualified a given reviewer is to review a particular paper; tractably finding good matchings; determining which papers can be rejected without full review; and identifying reviewers who have bid, reviewed, or scored papers in an attempt to manipulate outcomes. 

%
%
We addressed these challenges with a novel automated pipeline for reviewer--paper matching at AAAI 2021. Our pipeline consists of three main elements: (1) collecting and processing input data to identify problematic matches and generating reviewer--paper scores; (2) formulating and solving a constrained matching problem; and (3) splitting the reviewing process into two phases to better allocate reviewing resources towards borderline papers. We briefly describe each of these in turn (and then devote a full section to each in what follows). Code for our pipeline is available at \url{https://github.com/ChrisCameron1/LargeConferenceMatching}.

\newcommand{\minipar}[1]{\vspace{6pt plus 6pt}\noindent\textbf{#1.}~}
\minipar{Data Collection and Processing} Program Chairs need to transform a reviewer's profile into an estimate of how qualified that reviewer is to review each paper. Profiles can be as detailed as Program Chairs are creative, containing features such as
past email addresses,  DBLP/Semantic Scholar profiles, self-reported reviewing proficiency via a bidding phase, self-declared conflicts of interest (COIs), and texts of previously written papers. A key question that conference organizers need to answer is how to process this data and transform it into a numerical score assessing the quality of every plausible reviewer--paper pair. Two notable examples of such numerical scores are TPMS~\cite{charlin&al13} and ACL scores~\cite{neubig&al20}. 
Both of these scores are computed based on textual similarity between a submitted paper and reviewer's prior publications. However, they are not always available; TPMS scores can only be computed for reviewers whose prior publications are known and ACL scores can only be computed for reviewers whose past publication abstracts are accessible, e.g., via a known Semantic Scholar profile. 
In AAAI 2021, we computed TPMS and ACL scores for all applicable reviewer--paper pairs. We also knew 
reviewers' and papers'
primary and secondary subject areas and received explicit bids from reviewers. We aggregated all of our data to obtain final reviewer--paper scores that were between 0 and 1 and whose values were unbiased by missing data. We refer to the resulting scoring function as \emph{aggscore}. 

Input data is only as useful as it is trustworthy: a rising concern for large scale conferences is that 
participants
may self-report information to direct their submission to a specific reviewer or to ensure they review a specific paper
\cite{jecmen-etal-neurips2020}.
To that end, we constructed relationship graphs based on each 
user's
publication history to identify unreported conflicts (which, of course, also arose via mistakes or oversights from non-malicious users). We also developed a set of heuristics to identify suspicious patterns of self-reported conflicts and of bidding for papers to review. In both cases, we erred on the side of being cautious, forbidding all such suspicious assignments.

\minipar{Formulating a Constrained Matching Problem} Once scores have been assigned, the next key problem is to come up with an actual assignment. We formulated this problem as a Mixed-Integer Programming (MIP) problem. We are far from the first to use a MIP formulation for reviewer--paper matching~\cite{flach2010novel,garg&al10,tang&al10,lian&al18,taylor&al08,kobren&al19,charlin&al13,charlin&al11}. However, we went well beyond the vanilla ``find a score maximizing assignment'' formulation by including additional (soft) desiderata. For example, we penalized assigning coauthors to review the same paper, rewarded geographic diversity in assignments, and penalized assignments where reviewers each bid positively on each other's papers. These soft constraints served both to ensure that each paper was reviewed by an intellectually diverse set of reviewers and to make potential collusion rings (reviewers conspiring to review each other's papers) \cite{littman&al21} harder to sustain.
Separately, we introduced a seniority constraint to encourage our matching algorithm to distribute experienced reviewers across papers. 
We show in our experimental analysis that despite the mentioned advantages, these soft constraints only marginally impacted the mean aggregate score, and hence did not significantly degrade average match quality.

\begin{wrapfigure}{r}{0.6\textwidth} 
\centering
\includegraphics[width=0.5\textwidth]{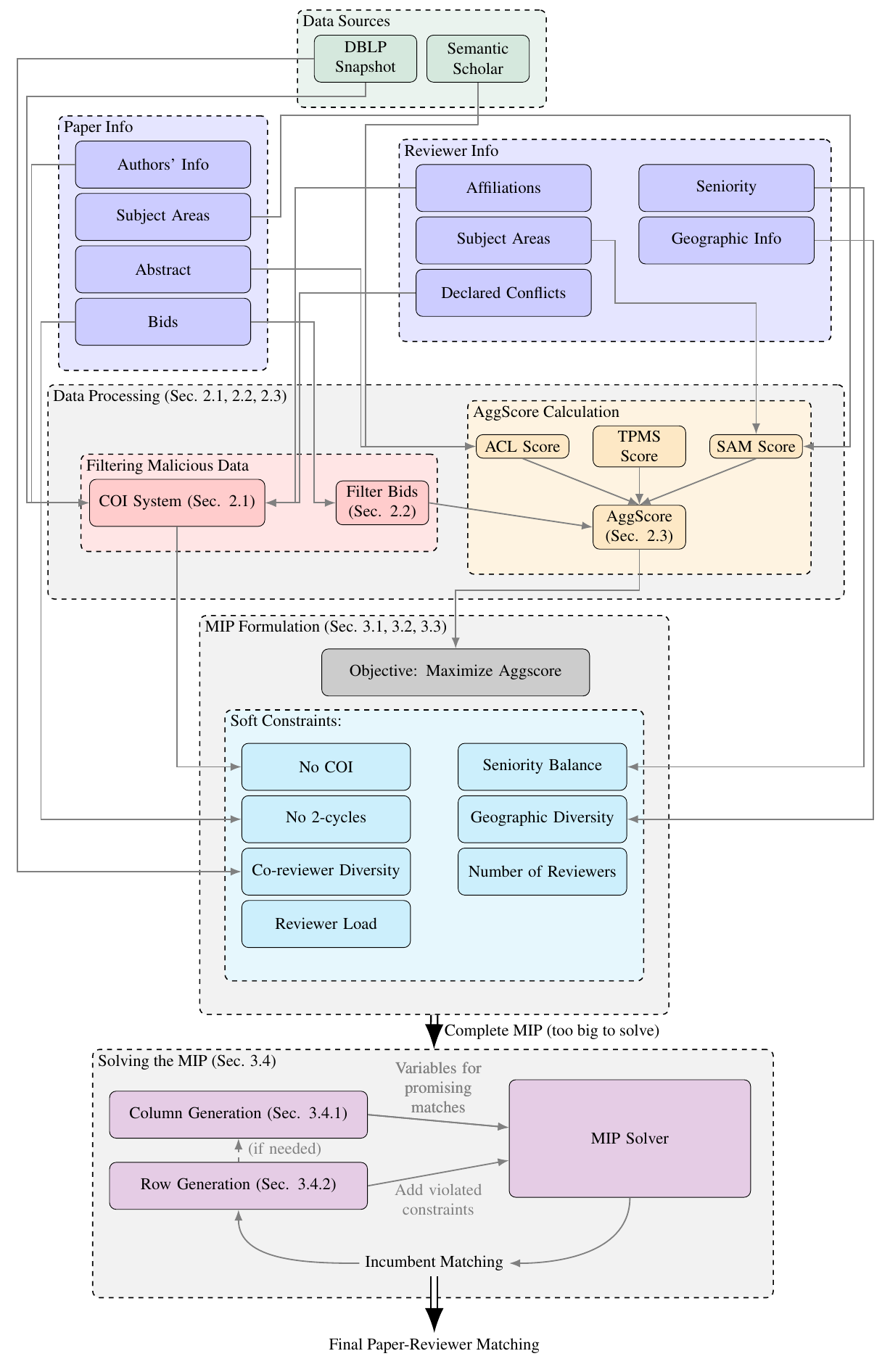}
\caption{Schematic diagram showing the flow of information in our system. 
}
\label{fig:flow_diagram}
\end{wrapfigure}

To ensure that the resulting MIP formulation could be solved within a reasonable amount of time, we sparsified our space of possible assignments by creating variables representing the possibility of assigning a given paper to a given reviewer only when the pairing had a sufficiently high score; we then used column generation as necessary to identify further candidate reviewers for hard-to-match papers. Our coauthorship constraints were too large to specify in memory, and so we similarly leveraged row generation (i.e., we identified an initial solution without coauthorship constraints and then iteratively added individual violated constraints and reran the optimization). See~\Cref{fig:flow_diagram} for how the information flows through our reviewer assignment system.

\minipar{Two-Phase Reviewer Assignment} The quality of papers submitted to conferences like AAAI 2021 varies vastly. Reviews are signals of quality, but can be very noisy. This was perhaps most convincingly demonstrated via an influential NeurIPS experiment \citep{lawrence&al17}, which split reviewing between two independent Program Committees. The experiment found that while the strongest and weakest papers could be reliably identified, many papers close to the decision boundary were accepted by one Program Committee and rejected by the other. This finding invites the idea of obtaining additional reviews for papers close to the decision boundary to reduce variance, and points out that one way to obtain this additional reviewing capacity is to reduce the reviewing load allocated to papers far below threshold.

One commonly used approach in journals which has also been adopted by recent AI conferences such as IJCAI 2020~\cite{bessiere&al20} summarily rejects some fraction of papers without full reviews (based, e.g., on quick perusal of each paper by one or more Area Chairs). However, this approach can be difficult to implement fairly (Area Chairs may not have strong opinions after a quick perusal; some papers will be missed; and such an approach would still impose a huge workload across 9,000 submissions). Even when it works well, the approach still leads to poorly explained decisions that many authors find frustrating (a quick perusal does not result in a detailed justification for the decision or suggestions about how work can be improved). 

We used an alternative to summary rejection, which we refer to as \textit{Two-Phase Reviewing}. This approach has various historical antecedents; we defer a survey of those to Section~\ref{sec: two-phase}. Roughly, instead of assigning, say, three reviews to each paper, we initially only assign two. If the two reviews agree that a paper should be rejected, the paper is indeed rejected without further reviews. A second phase of review assignments allocate two additional reviews to the papers that remain.  This approach redirects reviews to borderline papers while also giving meaningful feedback (two full reviews) to authors of rejected Phase 1 papers, and avoids the overhead of a separate summary rejection phase by treating all papers in the same way. It also has various other benefits. Notably, it gives Program Chairs a second opportunity to assign additional reviewers for papers identified as problematic or for which Phase 1 reviewers went missing; and also facilitates mechanisms like AAAI's ``NeurIPS/EMNLP FastTrack", in which high-scoring rejected papers from sister conferences with late notification dates are allowed to proceed directly to Phase 2, with their reviews from the sister conference taking the place of Phase 1 reviews.

We also describe the impact of these for AAAI 2021.
2,615 
papers (37\%)\footnote{Out of 9071 abstract submissions in Phase 1, only 7133 were accompanied by a full paper, out of which 37\% were rejected in Phase 1.} received two high-confidence reviews recommending rejection and so were rejected in Phase 1, which gave us a surplus of 2,615 
reviews (relative to AAAI's previous practice of assigning 3 reviews to every paper) to spread among the remaining papers in Phase 2 and FastTrack submissions. In AAAI 2021, the FastTrack received 778 abstract submissions and 688 full paper submissions, out of which 268 were ultimately accepted.
A key question is whether the rejected papers in Phase 1 would ultimately have been accepted if they received four full reviews. We devote Section~\ref{sec: evaluations} to an extensive real-data analysis that investigates such questions, but for this one (examined in detail in Section~\ref{sec:false-negatives}) we give a brief synopsis here. We identified an (inadvertent) natural experiment: when papers did not receive two full, high-confidence reviews in Phase 1, we promoted them to Phase 2 regardless of their scores and assigned additional reviewers. By subsampling pairs of reviews from the larger set that these papers eventually received, we estimate that there was only a 2.9\% probability that a Phase 1 rejected paper would eventually have been accepted to AAAI 2021. 


%


The remainder of this paper is organized as follows. We begin by describing the details of our approach for data collection and processing in \Cref{sec: input}. Then, we present our mixed-integer programming formulation of the reviewer--paper matching problem 
in \Cref{sec: MIP}. We describe our Two-Phase reviewing scheme 
in \Cref{sec: two-phase}. \Cref{sec: evaluations} consists of an exhaustive experimental analysis of data from AAAI 2021 in which we deployed these methods. Finally, we summarize our contributions in \Cref{sec: conclusion}. We note that much related work has studied different aspects of the reviewer--paper matching problem. To streamline exposition, we discuss related work in each corresponding section.

\section{Data collection and processing}
\label{sec: input}


In this section, we describe the techniques we designed for collecting and processing raw data about reviewers and papers into an aggregated score for reviewer--paper matching.

\subsection{Calculating Conflicts of Interests} 

A conflict of interest (COI) exists between a reviewer and a paper if the reviewer has a relationship with one or more of the paper's authors that gives them a personal interest in the paper's acceptance or rejection.
For example, most of us would struggle to provide a fully unbiased opinion about a paper authored by a former supervisor. Clearly, reviewers should not be matched to  papers with which they have a COI. 

\subsubsection{Filtering Manipulative Conflicts}
Existing reviewer--paper matching systems detect COIs by asking authors and reviewers to disclose all of their current and past affiliations as well as to declare any additional explicit conflicts.
This creates the possibility for a malicious author to provide very long lists of potential conflicts, either in an attempt to lead the system to choose a desired reviewer or simply to avoid expert assessment of their work.
We identified COI declarations as suspicious if they satisfied any of the following criteria: (1) a user listed an unusually large number of email domains as affiliations (8 or more); (2) a user listed an unusually large number of non co-authors as COIs (15 or more); (3) a user self-reported an unusually large number of COIs for which the corresponding reported users did not self-report a reciprocal COI (10 or more). 
We manually examined all such cases and removed all self-declared COIs that did not appear justified on inspecting the user's publicly available profile.
The exact parameters listed above were chosen empirically to capture only extreme behavior: the three conditions respectively captured only (1) 1.1\% of  users who reported more than a one conflict domain; (2) 0.14\% of users who self-reported at least one conflict; and (3) 0.44\% of users with at least one asymmetric conflict. In some cases, we observed evidence of quite egregious behavior: one user reported 35 conflict domains; another reported 57 asymmetric conflicts.  

\subsubsection{Predicting Unreported Conflicts of Interest}
\label{sec: COI}
\label{sec:COI}
 
Many conference management systems ask users to self-report information such as email domain names for organizations with which the user is currently or was formerly associated. 
Unfortunately, such user-provided information is often incomplete. Users can purposefully withhold information; they can forget to include a past affiliation; and
even if they remember every one, they can mistype a domain or fail to report all of their organization's aliases (\eg, \textit{iitd.ac.in} vs \textit{cse.iitd.ac.in}). 
We performed domain normalization and data cleaning (\eg, identifying common domain prefixes and suffixes, like \textit{.cse, .edu}, and \textit{.ac}, and unifying different aliases of the same organization), but obviously it cannot detect unreported or incorrect information.

We thus augmented user-reported conflicts with additional conflicts based on coauthorship information. We required authors and reviewers to supply DBLP profiles and downloaded the full DBLP database,\footnote{The full DBLP database is publicly available at \textit{\url{http://dblp.uni-trier.de/xml/dblp.xml.gz}}.} allowing us to identify coauthorship relations without the need to disambiguate similar names. We augmented this database with the current conference submissions and built a weighted coauthorship graph, with nodes corresponding to authors, edges existing between pairs of coauthors, and weights representing the number of publications coauthored by the pair.

One way to prevent COIs would be to prohibit a reviewer from reviewing a paper submitted by a previous coauthor. However, we considered this treatment too strict as it entirely forbids cross reviewing between authors who may not have interacted in years. (Instead, in Section~\ref{sec: constraining_matches}, we describe softer ways in which we discouraged our algorithm from selecting such matchings.) We thus considered a reviewer $r_j$ to have a COI with paper $p_i$ if any of the following conditions were satisfied for some author $a$ of $p_i$: 
\begin{enumerate}
\item $r_j$ coauthored a current conference submission with $a$; 
\item $r_j$ coauthored a published paper with $a$ over the previous 5 years; or 
\item $r_j$ and $a$ coauthored more than 6 published papers over any time period. 
\end{enumerate}
We further added conflicts between reviewer $r_j$ and papers by an author $a$ when we inferred that $a$ supervised $r_j$ or that both $r_j$ and $a$ were students of the same supervisor. We note that advisor--advisee relationships between users are unknown; hence we need to extract them from the coauthorship graph.
We inferred that $a$ was one of $r_j$'s supervisors if all of the following conditions held:
\begin{enumerate}
\item $a$ began publishing considerably earlier than $r_j$ (at least 5 years);
\item $a$ published at least 10 more papers than $r_j$; and
\item $a$ coauthored many of $r_j$'s early papers (at least 3 out of the first 10). 
\end{enumerate}

These inferences did not need to be perfect: each COI simply prevented a particular reviewer--paper matching, and a few false positive COIs were unlikely to be harmful. Indeed, as we show later in \Cref{fig:qualified} (\Cref{sec: scoring_function_evals}), most papers had a large number of highly qualified reviewers, indicating that occasionally forbidding one of them was unlikely to materially affect the quality of the matching.


We observe that we are not the first to use DBLP to infer COIs; \citet{long&al13} describe methods for inferring coauthor, colleague, advisor--advisee, and competitor relationships. Like us, they used DBLP data to identify coauthor relationships. However, they relied on users' public web pages to identify colleague and advisor--advisee relationships. We did not consider the latter approach feasible given the large scale of AAAI 2021. 


\subsection{Filtering Manipulative Bids}\label{sec:filter-manipulative-bids} 
Bids carry important information about reviewers' preferences that can significantly affect reviewers' satisfaction with an assignment. Unfortunately, reviewers can also maliciously submit bids with the aim of affecting a paper's chances of being accepted or rejected \cite{noothigattu2021loss, wu2021making}. 
Most past conferences have tried to detect such misbehavior via Area Chair (AC) oversight, spot checking of discussions, and whistle-blower reports. However, such approaches are labor intensive and scale poorly to huge conferences.

A recent paper \cite{jecmen-etal-neurips2020} proposed alleviating this issue by using a randomized algorithm for the reviewer assignment problem, decreasing the impact of malicious behavior by controlling the probability that certain (suspicious) reviewers or pairs of reviewers are matched to a paper.
However, such randomization can impose significant costs in terms of match quality.
We nevertheless strongly agree with \citeauthor{jecmen-etal-neurips2020} that malicious behavior is better prevented at the matching phase than detected afterwards. 
We found that the vast majority of papers at AAAI 2021 received a large number of bids by qualified reviewers. As was the case for COIs, the abundance of qualified reviewers allowed us to relatively harmlessly suppress a few potentially suspicious matches even in the likely event that many of them were false positives. 
At AAAI 2021, bidders could express their level of enthusiasm as \textit{not willing}, \textit{in a pinch}, \textit{willing}, and \textit{eager}.
To reduce the risk of making problematic assignments, we downgraded or suppressed a reviewer's bids when they appeared suspicious, which could happen in a number of ways.

First, a reviewer might place only a few bids to manipulate the likelihood that they would be assigned particular papers (in the most extreme example, bidding positively on only a single paper). To mitigate this risk, we discarded \emph{all} of a Program Committee (PC) member's bidding information if they placed fewer than $9$ positive bids overall (counting \emph{in-a-pinch} bids as half). 
We did something similar at other levels of enthusiasm: if a bidder placed fewer than $4$ \emph{eager} bids, we transformed them into \emph{willing} bids; then we transformed fewer than $4$ \emph{willing bids} into \emph{in-a-pinch}. We performed a similar procedure for Senior Program Committee (SPC) members, setting 
a higher threshold of $10$
to reflect the higher number of required positive bids.

Second, a reviewer could increase their chances of being assigned a particular paper by bidding negatively on many other papers for which they were qualified. We therefore discarded all \emph{not willing} bids of PCs and SPCs if the number of \emph{not willing} bids exceeded six times the number of \emph{willing} and \emph{eager} bids. 

Third, and most complex, \emph{collusion rings} are arrangements under which two or more reviewers agree to positively review each other's papers, a dismaying phenomenon that is reportedly on the rise \cite{littman&al21}. We define two criteria to judge whether a bid by a reviewer $r$ on a paper $p$ is consistent with participation in a collusion ring: (1) at least $40$\% of $r$'s positive bids are on papers authored by 
a specific
author of $p$; (2) 
there exists some author $r^*$ of $p$ such that both $r$ and $r^*$ bid positively on each other's papers. Specifically, the sum of the number of $r$'s positive bids on papers authored by $r^*$ and the number of $r^*$'s positive bids on papers authored by $r$ is at least $60$\% of 
the total number of positive bids by $r$ and $r^*$. 
For both (1) and (2) above, we considered \emph{eager} and \emph{willing} bids as positive.
These criteria are not perfect; e.g., (1) can occur in small subcommunities where a few researchers focus on a given topic, and (2) can wrongly flag reviewers with a few submissions (e.g., two authors with one paper each bid on each other's paper). 
We were willing to tolerate some false positives (for the same reasons outlined above), but aimed to reduce their number by requiring a substantial pattern of suspicious bidding across all of a reviewer's bids. In the assignment process we discarded \emph{all} of a reviewer's bids when 
 either (1) or (2) indicate the reviewer's participation in a collusion ring.

We conducted an extensive manual examination of bids, using various heuristic criteria. In all cases that we judged problematic, our method had already eliminated the appropriate bids.




\label{sec: bids_processing}

\subsection{Scoring Matches}
\label{sec: scoring_function}

A key question in automated reviewer--paper matching is quantifying the value of matching a reviewer $r_j$ with a submitted paper $p_i$, which we denote as $\scorematrixentry_{ij}$.
It is critical to get this right.\footnote{Indeed, a few readers may remember that our initial matching at AAAI 2021 had to be changed, and the reason was that we accidentally miscalibrated our scoring function. Based on the emails we received, reviewers definitely noticed!}
In brief, we considered a match to be good if the reviewer both had expertise in the paper's subject matter and was interested in the paper. We assessed expertise by aggregating three complementary signals: Toronto Paper Matching System (TPMS) scores; ACL matching scores; and degree of match between the paper's primary and secondary subject area keywords and those of the reviewers. We assessed degree of interest via bids, assuming a baseline level of interest for papers about which a reviewer was predicted to have expertise and for which the reviewer did not submit any explicit (positive or negative) bid.

Much existing work has leveraged similar signals, much of which is focused on automatically discovering latent topics of a submission and comparing it with latent topics of a set of papers authored by a given reviewer \citep{mimno&al07, long&al13, anjum&al19, kobren&al19, wieting&al19}.
Three further lines of related work deserve detailed discussion, since we either extended them \cite{conry&al09} or leveraged them directly \cite{charlin&al13,neubig&al20}. 

First, \citet{conry&al09} proposed a score based on the primary and secondary subject areas of the paper and that of the reviewer. The fundamental idea was to embed both reviewers and papers into a discrete vector space 
with each discrete dimension quantifying the affinity of either the reviewer or the paper to the corresponding topic. 

Second, the TPMS score  \cite{charlin&al13} is computed based on the similarity between the text of a submitted paper and a reviewer's prior publications. 
The similarity is quantified by the dot product between a paper vector and a reviewer vector. The paper vector is constructed using the topic proportions in the text of the submitted paper identified by Latent Dirichlet Allocation (LDA) \cite{blei&al03}. The reviewer vector is computed by taking an average of the topic proportions in each of the reviewer's prior publications.
TPMS scores can only be computed for reviewers whose prior publications are known. 

Third, the ACL matching score \cite{neubig&al20} compares abstracts of a submitted paper and the abstracts of a reviewer's prior publications in a dense vector space.
The abstract representation is computed by averaging the embeddings of subwords generated by the SentencePiece tokenizer \cite{kudo-richardson-2018-sentencepiece}. 
 A score is then computed for each of the reviewer's prior publications with respect to the paper. The top three scores are aggregated to compute the ACL score. 
ACL scores can only be computed for reviewers whose past publication abstracts are accessible, e.g., via a known Semantic Scholar profile. 

The rest of this section is organized as follows. We first describe how we computed Subject Area Matching (SAM) scores, a continuous extension of discrete affinity scores. Then, we describe how we normalized raw TPMS, ACL and SAM scores to make sure that they have the same scale. We then explain how we merged these scores to make them robust to missing data. This was a real concern because we needed to rely on reviewers to manually provide the information required to compute these scores, and around 40\% of the reviewers had at least once missing entry. Next, we explain how we adjusted these scores to take into account reviewers' bids. Finally, we describe an approach for pruning low-scoring matches, which we used to avoid the possibility of wildly inappropriate matches.

\subsubsection{Subject Area Matching Score (SAM)}
\label{subsec:SAMS}
We scored the keyword match between papers and reviewers via an extension of the affinity scoring idea described earlier \cite{conry&al09}, which embeds each reviewer and paper in a 
vector space. Our approach went beyond explicitly specified subject areas by learning correlations between different keywords. This is important because the approach of \citet{conry&al09} assigns a weight of 0 to keywords that are not specified explicitly, even when these have a strong degree of semantic similarity with a keyword that \textit{is} specified. 
For example, keywords \textit{``Humans and AI -> Human-Agent Negotiation"} and \textit{``Game Theory and Economic Paradigms -> Negotiation and Contract-Based Systems"} were highly correlated at AAAI 2021, and thus it might be inaccurate to assign a weight of $1$ to only one of the two when the other is absent.

In AAAI 2021, authors described their submissions via one primary keyword and a variable number of secondary keywords; likewise reviewers characterized their area(s) of expertise. Let $\subjectareas=\{\subjectarea_1, \cdots \subjectarea_{N_\subjectarea}\}$ be the set of subject areas. We represent each paper $p_i$ using two vectors based on their associated subject areas: a primary vector $\textbf{i}_p \in \{0,1\}^{N_\subjectarea}$ and secondary vector $\textbf{i}_s \in \{0,1\}^{N_\subjectarea}$. $\textbf{i}_p$ is a one hot vector that represents the primary subject area and $\textbf{i}_p$ is a multi hot vector that represents the secondary subject areas. We represent each reviewer $r_j$ using a vector $\textbf{j} \in [0,1]^{N_\subjectarea}$. We set the dimensions corresponding to the reviewer's primary subject area to 1 and the reviewer's secondary subject areas to 0.5.
Constructing a sparse vector representation using just the primary and secondary areas will fail to capture reviewer's expertise in semantically similar areas that were not explicitly mentioned. To alleviate this problem, we convert the sparse $\textbf{j}$ vector to a dense vector by inferring the reviewer's expertise in subject areas other than primary and secondary.
 
Let $\subjectarea^\prime \in \subjectareas$ be a subject area which is neither in reviewer's primary or secondary areas. We first identify multiple sources that can help us infer the expertise value $j_{\subjectarea^\prime} \in [0,1]$ for $\subjectarea^\prime$. We then infer an expertise value using each source one at a time and aggregate them. We now discuss each source and how we infer the expertise value using them.

\textit{Co-occurrence:} Given that a reviewer is an expert in $\psi$, we compute conditional expertise in $\psi^\prime$  as $P_e(\psi^\prime|\psi)=n(\psi,\psi^\prime)/n(\psi)$, where $n(\psi,\psi^\prime)$ is the number of reviewers and papers with both $\psi$ and $\psi^\prime$ in their subject areas and $n(\psi)$ is the number of reviewers and papers with $\psi$ in their subject area. The intuition behind this approximation is two fold: (1) $P_e$ uses the current trend to capture correlation among the subject areas and (2) the asymmetry in $P_e$ ensures expertise in a specialized area can indicate expertise in that generic areas but not vice-versa. Given a reviewer's subject area $\psi$, the expertise value of $\subjectarea^\prime$ is $\rho * P_e(\psi^\prime|\psi)$. $\rho$ is set to 1 when $\psi$ is a primary area and 0.5 when $\psi$ is a secondary area. We consider the primary and the secondary areas as different sources and compute a expertise value for each source separately. 

\textit{Paper Subject Areas:} The subject areas of papers submitted by $j$ (if any) that are not already in the reviewer's subject areas are set to a value (0.2) lower than the one used for secondary areas. This helps capture expertise of reviewers in areas they just started exploring.

\textit{Common Parent:} Each subject area $\subjectarea$ can be expressed as $\textit{parent}(\subjectarea) \rightarrow \textit{child}(\subjectarea)$. For example, in \textit{NLP} $\rightarrow$ \textit{Dialog Systems}, \textit{NLP} is the parent and \textit{Dialog Systems} is the child. Each parent can have multiple children. Given a reviewer's subject area $\psi$, we assign a score of $\rho * 0.4$ to $\subjectarea^\prime$ if $\textit{parent}(\subjectarea^\prime)$ is same as $\textit{parent}(\subjectarea)$ and 0 otherwise. $\rho$ is same as the one used for co-occurrence computation.

For each subject area $\subjectarea^\prime$, we compute the expertise value from different sources, aggregate them using the max operator and assign it to $j_{\subjectarea^\prime}$. The max operator helps pick the source that best models $\subjectarea^\prime$. Finally, we compute the \emph{Subject Area Matching (SAM)} score  using sparse paper vectors ($\textbf{i}_p$ and $\textbf{i}_s$) for paper $p_i$ and dense reviewer vector \textbf{j} for reviewer $r_j$ as 
$
    \text{SAM}(i, j) = \frac{\textbf{i}_p \cdot \textbf{j} + \boldsymbol{\lambda}  \cdot \textit{Sorted}(\textbf{i}_s \circ \textbf{j})}{Z},
$
\noindent where the operator $\circ$ indicates Hadamard product, the function \textit{Sorted}($\textbf{x}$) sorts the elements in $\textbf{x}$ in decreasing order and  $\boldsymbol{\lambda} = [0.5, 0.5^2, \cdots, 0.5^{N_\subjectarea}]$. The normalizer $Z=1+\sum_{m=1}^{M}{0.5}^m$, where $M$ is the number of non-zero values in $\textbf{i}_s$. The scoring function ensures (1) paper's primary area contributes more than all the secondary areas, (2) the more the number of secondary areas match, the higher the score, and (3) papers do not get penalized for providing fewer (or zero) secondary areas.


Some intuitive properties of SAM scores are: it considers the primary keyword to be more important than all of the secondary keywords; gives each individual keyword less weight when more keywords are listed; and imputes additional keywords (with lower weights) for reviewers based on keyword co-occurrence patterns across the whole conference. 







\subsubsection{Normalizing Scores}
We wanted all final scores to lie in the range $[0,1]$ so that we could use consistent units to set the parameters that penalize violations of soft constraints (e.g., we might penalize violation of a soft constraint by $0.1$ units, meaning that we would accept a match that scored $0.09$ units worse in order to preserve the constraint, but not a match that scored $0.11$ units worse).
Since we computed the three component scores (TPMS, ACL and SAM) independently, and since we observed a huge discrepancy between the numerical range of the three raw scores, we first normalized them each individually before taking their weighted average as the final matching score.

%
To do so, for each score, we independently and randomly selected many (paper, reviewer) pairs. A Program Chair manually examined each pair (reading the paper's abstract and the reviewer's profiles) and annotated each pair $(p_i,r_j)$ as $y_{ij} \in \{1, 0.75, 0.5, 0\}$, corresponding to \textit{excellent}, \textit{good}, \textit{in-a-pinch}, and \textit{bad} matches respectively. We then performed a linear regression mapping the raw score component to $y_{ij}$ and clipped the resulting function to the range $[0,1]$.

\subsubsection{Aggregating Scores} 
\label{subsec:scoreagg}
Our next task was to aggregate scores into a single number between $0$ and $1$, in a way that was robust to missing TPMS and ACL scores. We decided to give more weight to keyword-based scores than either of the automatically inferred scores (TPMS and ACL), since keyword scores were computed using information provided explicitly by authors and reviewers. 
When both TPMS and ACL were present, we summed TPMS + ACL + 2 * SAM and then renormalized; when only one of TPMS and ACL were present, we summed the 
 available
score and SAM score and renormalized; when only SAM score was present we used it as the base aggregated score:
\begin{equation*}
\text{Base Aggregated Score} =\begin{cases} \frac{1}{4}\text{TPMS} + \frac{1}{4}\text{ACL} + \frac{1}{2}\text{SAM} & \text{if all data is present} \\
\frac{1}{2}\text{ACL} + \frac{1}{2}\text{SAM} & \text{if TPMS is missing} \\
\frac{1}{2}\text{TPMS} + \frac{1}{2}\text{SAM} & \text{if ACL is missing} \\
 \text{SAM} & \text{if both ACL and TPMS are missing}
\end{cases}
\end{equation*}



\subsubsection{Accounting for bids} Given this score capturing reviewer \emph{expertise}, we still needed to account for reviewer \emph{interest} as expressed through bidding to obtain a final result we called \emph{aggscore}. We computed aggscore by upweighting/downweighting the base aggregated score based on the bid. Our approach to combine expertise and interests does not allow bids to override the base aggregated scores. We achieved this by computing aggscore as follows:
\begin{align*}
    \rm{aggscore}=(\rm{Base~aggregated~score})^{\rm{bidscore}}
\end{align*}
We used bidscores as $20, 1, 0.67, 0.4, 0.25$ corresponding to \textit{not-willing, not-entered, in-a-pinch, willing, and eager} bids, respectively.
Observe that this procedure maintained scores in the range between $0$ and $1$; exponents greater than $1$ (for \emph{not-willing}) reduced scores while exponents less than $1$ (for \emph{in-a-pinch, willing,} and \emph{eager}) increased scores. Our use of an exponent of $1$ for papers that received no explicit bid left those scores unchanged.
Observe also that our use of exponentiation (as compared e.g., to adding a constant and renormalizing) meant that reviewers could not influence the system to give them papers for which they were manifestly unqualified, whereas the presence or absence of a bid could make a significant difference to the scores of sufficiently qualified reviewers.

\subsubsection{Cleaning up low scoring matches} Finally, we observed that while high ACL or TPMS scores indicated good matches, their assignment of low values carried much less signal than low SAM scores. Hence, when aggscore was less than $0.15$, we used the minimum of $0.15$ and the SAM score exponentiated by the reviewer's bid.
This transformation ensured that whenever a low-scoring match was selected, it was at least from the same subject area. This avoided the possibility of wildly inappropriate matches and helped reviewers see some rationale for our decisions to assign poorly matching papers.

\section{Formulating the Review Assignment problem}
\label{sec: MIP}
We formulated the problem of maximizing our scoring function subject to our set of both hard and soft constraints as a Mixed-Integer Programming (MIP) problem. 
It is quite natural to formulate the
reviewer--paper matching problem as one of maximizing an aggregate weighted matching score subject to reviewer and paper capacity constraints and avoiding COIs, and much past work has done so  \cite{flach2010novel,garg&al10,tang&al10,lian&al18,taylor&al08,kobren&al19,charlin&al13,charlin&al11}.  
Indeed, with only COI and capacity constraints, this formulation corresponds to a network flow problem and is thus solvable in polynomial time~\cite{taylor&al08}. 
Unfortunately, that formulation can yield an inequitable distribution of work across reviewers and of scores across papers.
To address this, \citet{stelmakh2019peerreview4all} and \citet{kobren&al19} focused on finding more balanced matchings.  Specifically, instead of the common objective of maximizing the aggregate weighted matching score,  \citet{stelmakh2019peerreview4all} look into finding matches that maximize the minimum matching score across all papers. Because the resulting optimization problem is NP-hard, they design an approximation algorithm, which finds a suboptimal solution by solving a series of network flow problems. \citeauthor{kobren&al19}---like us---aim to maximize an aggregate weighted matching score, but took a somewhat different approach. Specifically, they augmented an Integer Linear Program with lower and upper bounds on the assignments per reviewer and also added local fairness constraints to ensure that the total matching score for each paper would exceed a threshold; they propose an approximation algorithm that relaxes integrality constraints to solve their assignment problem. Our own approach uses slack variables to soften hard constraints instead of relaxing the integer constraints, resulting in a mixed-integer program. 
We also introduce a wide range of additional constraints. 
We now describe our approach in more detail.

\subsection{Constraining Matches}
\label{sec: constraining_matches}

It would not suffice simply to identify a score-maximizing matching: the best reviewers could be given hundreds of papers; different papers could receive very different treatment, leading to an unfair assignment. For example, a paper may get assigned all junior reviewers from one research group. 
We thus imposed various \emph{constraints} on reviewer--paper matchings.

\newtheorem{constraint}{Constraint}

\begin{constraint}[COI] 
\label{cons:coi}
No reviewer should be assigned to a paper with which they are in conflict.\end{constraint}

\begin{constraint}[Number of Reviewers]
\label{cons:numreviewers}
Each paper must receive a specified number of reviews. 
\end{constraint}
Historically, this number of reviews has been three at AAAI and many other AI conferences. In our  two-phase reviewing system (see Section~\ref{sec: two-phase}) we solved the reviewer--paper matching problem separately in each of our phases, assigning two reviewers for each paper in Phase 1, 
and adding additional reviewers in Phase 2 such that each paper not rejected in Phase 1 got at least four reviews across the two phases. (That is, we added extra reviewers in Phase 2 when reviewers didn't submit their reviews at the end of Phase 1.)

\begin{constraint}[Reviewer Load]
\label{cons:reviewerload}
The number of papers assigned to any reviewer should not exceed some maximum. 
\end{constraint}
At AAAI 2021 we had a hard limit of three papers per reviewer per phase, and (as we will discuss later) penalized matchings that assigned more than two papers per reviewer in a given phase.



\begin{constraint}[Seniority]
\label{cons:seniority}
Each paper is assigned at least one experienced reviewer. 
\end{constraint}
Our goals were: (1) to ensure that experienced reviewers were distributed fairly across papers; (2) to reduce variance in reviews; and (3) to ensure that each discussion had at least one experienced participant, to promote the training of novice reviewers. 

A second set of constraints served both to ensure that each paper is reviewed by an intellectually diverse set of reviewers and to make collusion rings harder to sustain. Collusion is only possible amongst reviewers who can somehow communicate with one another. By preventing pairs of co-authors from reviewing the same papers,  preventing arbitrary pairs of reviewers from bidding to review each other's papers, and by choosing reviewers from as diverse a set of geographic regions as possible, we limited opportunities for gaming the system.

\begin{constraint}[Coauthorship Distance] 
\label{cons:coauthor}
No two reviewers assigned to the same paper have small distance in the coauthorship graph.
\end{constraint}

\begin{constraint}[Geographic Diversity]
\label{cons:geo}
No two reviewers assigned to the same paper belong to the same geographic region. 
\end{constraint}

\begin{constraint}[No 2-Cycles]
\label{cons:cycle}
No pair of reviewers who both bid positively on each other's papers may review each other's submissions. 
\end{constraint}

\vspace{.3em}
It would not have been desirable to satisfy these constraints at all costs. 
For many papers, it was easy to find qualified reviewers satisfying all of these constraints; however, for others imposing all of these constraints would dramatically have degraded reviewer quality or even made it impossible to find the required number of reviewers.
For example, qualified reviewers often bid on each other's papers without harboring malicious intent. Out of caution, we adopted the principle of finding matchings that avoided such cases when reasonable alternatives were present, but were willing to relax the principle otherwise. We therefore expressed all but \textit{COI},  \textit{Number of Reviewers}, and PC's \textit{Reviewer Load} as soft constraints. More specifically, for each constraint we identified a constant expressing its importance and penalized the objective function by this constant each time the constraint was violated for a reviewer--paper pair. Observe that the task of identifying these constants was dramatically simplified by our decision to normalize our scoring function between $0$ (worst possible reviewer--paper matching) and $1$ (best possible matching). For example, expressing a coauthorship distance penalty of $0.1$ would mean that we would prefer to accept a reviewer--paper matching scoring up to $0.1$ points lower in exchange for satisfying the constraint.

\newcommand{\constraintpar}[1]{\vspace{.6em}\noindent\textbf{#1}}

\subsection{MIP Problem Formulation}
\label{subsec: MIP_def}

 

We now formally specify our mixed-integer program. 
Let us denote the set of $N$ papers by $\papers$, the set of Program Committee members (PCs) by $\pcs$, the set of SPCs (Senior Program Committee members) by $\spcs$, and the set of Area Chairs (ACs) by $\acs$. We define the set of reviewers to be $\reviewers = \pcs \cup \spcs \cup \acs$ and use the term \textit{reviewer} to refer to any of a PC, SPC or AC.

 
 Our goal is to assign $\gamma_\text{pc}$ PCs, $\gamma_\text{spc}$ SPCs, and $\gamma_\text{ac}$ ACs to each paper $p_i$, for $1 \leq i \leq N$, such that the total reviewer--paper matching score is maximized subject to a set of constraints denoted by $\constraints$.
Our overall objective function not only includes the total matching score,
but also various penalties via the slack variables, due to violation of corresponding hard constraints.
We initialize the overall objective function $\obj$ as the total matching score and then keep on adding terms to it as we relax the various constraints in the next section. 
We create binary variables $\assignmatentry_{ij}$ that denote whether paper $p_i$ is assigned to reviewer $r_j$.\footnote{In Phase 2, we set $\assignmatentry_{ij} = 1$ if $r_j$ was already assigned to $p_i$ in Phase 1 to ensure that the matching from Phase 2 is a superset of Phase 1 matching.
The capacity limits on reviewers were increased accordingly.}
 We can write:
\begin{align}
\label{eq:matchscore}
\text{Initial Objective:~~}  \obj^\text{match} = \sum\limits_{i \in \papers, j \in \reviewers}
\scorematrixentry_{ij} \assignmatentry_{ij}.
\end{align}
Recall that $\scorematrixentry_{ij} \in [0,1]$ is the aggscore, the matching score between reviewer $r_j$ and paper $p_i$ as discussed in Section~\ref{sec: scoring_function}. Next, we formulate all the (soft) constraints in $\constraints$ as described earlier.

\subsection{Formulation of Constraints in $\constraints$}
\label{subsec: MIP_constraints}

\constraintpar{Constraint \ref{cons:coi}} (COI):
Let $\coimatentry_{ij} = 1$, if there exists a conflict of interest between reviewer $j$ and any of the authors of the paper $i$, and $0$ otherwise. 
Reviewer $j$ cannot review paper $i$ if $\coimatentry_{ij} = 1$:
\begin{equation}
    \assignmatentry_{ij} = 0, \text{\  for each \ } i \in \papers, j \in \reviewers, \text{ \ such that \ } \coimatentry_{ij} = 1.
\end{equation}


\constraintpar{Constraint \ref{cons:numreviewers}} (Number of reviewers):
For each paper $p_i$, we need to ensure that it gets the desired number of reviewers:

\noindent \begin{equation*}
    \label{eq:papercapacity}
    \sum\limits_{j \in \pcs} \assignmatentry_{ij} = \gamma_\text{pc} \ \ ; \ \ 
    \sum\limits_{j \in \spcs} \assignmatentry_{ij} = \gamma_\text{spc} \ \ ; \ \ 
    \sum\limits_{j \in \acs} \assignmatentry_{ij} = \gamma_\text{ac} 
\end{equation*}
With the equality constraints above, the MIP may become infeasible. Hence, we replace it with inequality constraints: 
\noindent \begin{equation}
    \label{eq:papercapacity}
    \sum\limits_{j \in \pcs} \assignmatentry_{ij} \leq \gamma_\text{pc} \ \ ; \ \ 
    \sum\limits_{j \in \spcs} \assignmatentry_{ij} \leq \gamma_\text{spc} \ \ ; \ \ 
    \sum\limits_{j \in \acs} \assignmatentry_{ij} \leq \gamma_\text{ac} 
\end{equation}
These constraints would be tight for a maximization problem where each match contributes positively to the objective. Note that it is possible to have a case in which conditional on the rest of the assignment, the only remaining matches do not positively contribute to the objective function. In such a case, the optimizer will not assign those papers their full capacity of reviewers since such assignments would be of poor quality.
We manually reviewed such cases.

\constraintpar{Constraint \ref{cons:reviewerload}} (Reviewer Load): The total number of papers assigned to a reviewer $j$ should not exceed their capacity $\capacity_j \in \mathbb{Z^+}$:
\begin{equation}
    \label{eq:reviewercapacity}
    \sum\limits_{i \in \papers} \assignmatentry_{ij} \leq \capacity_j, \forall j \in \pcs 
\end{equation}
\begin{equation*}
    \label{eq:reviewercapacity}
    \sum\limits_{i \in \papers} \assignmatentry_{ij} \leq \capacity_j, \forall j \in \spcs \cup \acs 
\end{equation*}
Just ensuring the above upper bound may lead to an inequitable assignment of papers to reviewers, especially ACs and SPCs, many of whom may end up with no assignment at all. 
To resolve this, one option is to add a lower bound $\lowerlimitcapacity_j$ on the number of papers assigned to ACs and SPCs, as suggested by \citet{kobren&al19}:
\begin{equation*}
    \label{eq:reviewerlowerlimit}
    \sum\limits_{i \in \papers} \assignmatentry_{ij} \geq \lowerlimitcapacity_j, \forall j \in \spcs \cup \acs.
\end{equation*}
However, the lower bound constraint only ensures that no SPC or AC is unassigned, rather than preventing skewed matchings. 
To promote an equitable distribution of papers amongst reviewers, we introduce $W$ intermediate capacity levels. For each $w \in \{1,2,\dots,W \}$, we define $\capacityentry_{jw}$
to be the intermediate capacity  for reviewer $j$ and set $\capacityentry_{j1}=\lowerlimitcapacity_j$ and
$\capacityentry_{jW} = \capacity_j$.
For ACs and SPCs, we replace the hard capacity constraints with a sequence of soft capacity constraints, one for each capacity level $w$. These constraints promote equitable distribution of papers as additional penalties are incurred each time an assignment crosses a new capacity level, making it very expensive to overload an individual.

To do so, we introduce new slack variables, $\capacityslack_{jw} \ge 0, \forall j \in \spcs \cup \acs, \forall w \in \{1 \ldots W\}$, each of which relax the capacity $\capacityentry_{jw}$ to  $\capacityslack_{jw} + \capacityentry_{jw}$, but penalize the objective by $\capacityslackpenalty_w(\text{type}_j)$ for each additional assignment above $\capacityentry_{jw}$, where $\text{type}_j$ denotes the reviewer category (SPC or AC). 
Then, for each $w \in \{1 \ldots W\}$:
\begin{align}
    \label{eq:reviewercapacityslack}
    \sum\limits_{i \in \papers} \assignmatentry_{ij} \leq & \  \capacityslack_{jw} + \capacityentry_{jw}, \ \  \forall j \in \spcs \cup \acs \text{ and } \forall w \in \{1 \ldots W\};  \\
    \label{eq:reviewercapacitypen}
    \obj^{\text{cap}} = &  \sum_{w \in W} \sum_{j \in \spcs \cup \acs} \capacityslackpenalty_w(\text{type}_j)\capacityslack_{jw}.
\end{align}


\constraintpar{Constraint \ref{cons:seniority}} (Seniority):
We first assign $\text{seniority}_j \in \{0, 1, 2, 3\}$ to each reviewer $j$, such that $3$ is the senior-most category and $0$ is the junior-most category. In AAAI 2021, we chose to assign seniority levels as follows: Reviewers who had previously reviewed for at least 3 relevant conferences\footnote{Reviewers were asked: ``How many times have you been part of the Program Committee (as PC/SPC/AC, etc) of AAAI, IJCAI, NeurIPS, ACL, SIGIR, WWW, RSS, NAACL, KDD, IROS, ICRA, ICML, ICCV, EMNLP, EC, CVPR, AAMAS, HCOMP, HRI, ICAPS, ICDM, ICLR, ICWSM, IUI, KR, SAT, WSDM, UAI, AISTATS, COLT, CORL, CP, CPAIOR, ECAI, or ECML in the past?''} or had published 10 or more papers in relevant conferences were assigned seniority level $3$. Remaining reviewers having between 4 and 9 published papers were assigned seniority level $2$. Remaining reviewers having 2 to 4 papers or with experience reviewing at one or two previous conferences were assigned seniority level $1$. All other reviewers were assigned a seniority level of $0$.

We then reward each paper for maximizing its seniority level, up to some maximum reward. We introduce  slack  variables, $\seniorityslack_i\geq 0 ~ \forall i \in \papers$.
\begin{align}
    \label{eq:seniorityconstraints}
    \seniorityslack_i \leq& \sum\limits_{j \in \pcs} \text{seniority}_j \assignmatentry_{ij} , \forall i \in \papers \\
    \seniorityslack_i \leq& \text{~TargetSeniority}, \forall i \in \papers  \\
    \seniorityslack_i \geq& \text{~MinSeniority} , \forall i \in \papers \\
    \label{eq:senioritypenalty}
    \obj^{\text{sen}} =& \sum\limits_{i \in \papers} \seniorityreward \seniorityslack_i 
\end{align}

Note that setting $\text{MinSeniority}$ greater than 0 can result in possible infeasibility (i.e., it may not be possible to assign a minimum level of seniority simultaneously to every paper). Setting $\text{MinSeniority}=0$ keeps the constraint soft. Papers are rewarded for seniority up to a maximum seniority level given by $\text{TargetSeniority}$. Setting this target allows us to express a preference for spreading seniority out over many papers rather than concentrating it on a few papers, as the objective only gets a bonus for the first $\text{TargetSeniority}$ units of seniority per paper. 


\constraintpar{Constraint \ref{cons:coauthor} }(Coauthorship Distance):
This set of constraints penalizes the assignment of  pairs of reviewers who have ever coauthored a paper together to the same paper. This reduces the chances of papers' reviewers knowing each other, promoting review diversity.
To do so, for each pair of reviewers in $\coreviewerset = \{(j,j') | j < j', j,j' \in \pcs \cup \spcs\}$, we define a new variable $\coreview_{jj'} \ge 0$ such that it takes the value of $1$ if there exists a common paper assigned to both $j$ and $j'$, and $0$ otherwise. 
We exclude ACs from this set of constraints as they are few in number, highly trusted, and are not directly involved in discussions with the PCs.
The following constraints define $\coreview_{jj'}$: 
\begin{align}
    \coreview_{jj'} \ge& \ 0, \forall j,j' \in \coreviewerset; \\
    \coreview_{jj'} \ge& \ \assignmatentry_{ij} + \assignmatentry_{ij'} - 1, \ \ \forall i \in \papers \text{ and } \forall j,j' \in \coreviewerset.
\end{align}
Next, let $\coauthordist_{jj'}$ denote the edge distance between two reviewers $j$ and $j'$ in a coauthorship graph with reviewers as nodes. 
An edge exists between $j$ and $j'$ iff they have ever published a paper together (based on DBLP data, as described in Section~\ref{sec:COI}).
To discourage coauthors being reviewers of the same paper, we add a penalty that is bigger when coauthorship distance is small: 
\begin{align}
    \obj^{\text{co}} = & \sum\limits_{(j,j') \in \coreviewerset} p^{co}(\coauthordist_{jj'})\coreview_{jj'}.
\end{align}
$p^{co}(d)$ is the penalty coefficient, which depends on coauthorship edge distance $d$. We set $|p^{co}(1)| > | p^{co}(2) | > 0$ and $p^{co}(d) = 0, \forall d \ge 3$.

\constraintpar{Constraint \ref{cons:geo} }(Geographic Diversity):
For each paper $i$, let variable $\numregion_i$  count the number of distinct geographic regions of its assigned PCs and SPC. 
We inferred regions based on academic affiliations and email addresses provided by the reviewers. 
To encourage assignment of reviewers from different geographic regions, we add a reward proportional to $\numregion_i$ in the overall objective function so that $\numregion_i$ is maximized.
To count $\numregion_i$, we need to introduce auxiliary indicator variables, $\paperregion_{i\region} \leq 1$, that take the value of $1$ only if paper $i$ is assigned at least $1$ reviewer from region $\region \in \regions$.
Let $\reviewerregion_{j\region}$ indicate if reviewer $j$ belongs to region $\region$ or not.
The following expressions define the overall regional constraints for each paper.
\begin{align}
    \label{eq:regionconstraints}
    \numregion_i \leq& \sum_{\region \in \regions} \paperregion_{i\region}, \ \forall i \in \papers \\
    \paperregion_{i\region} \leq& \sum\limits_{j \in \pcs \cup \spcs} \reviewerregion_{j\region} \assignmatentry_{ij}, \forall \region \in \regions \text{ and } \forall i \in \papers\\
    \paperregion_{i\region} \leq& \ 1,   \forall \region \in \regions \text{ and } \forall i \in \papers \\
    \label{eq:regionreward}
    \obj^{\text{reg}} =& \sum\limits_{i \in \papers} \regionreward \numregion_i 
\end{align}

\constraintpar{Constraint \ref{cons:cycle} }(No 2-Cycles):
These constraints aim to avoid scenarios where two reviewers both bid positively on each other's papers and are assigned these papers to review. Similar to Constraint \ref{cons:coauthor} (Coauthorship Distance) and \ref{cons:geo} (Geographic Diversity), we exclude ACs from this constraint for similar reasons.
We first identify all such bidding cycles, where reviewer $j$ bids positively on paper $i$ authored by reviewer $j'$ and $j'$ bids positively on paper $i'$ authored by reviewer $j$, with $j,j' \in \pcs \cup \spcs$.
Denote such a bidding cycle by a 4-tuple $(j,j',i,i')$, and the set of all such cycles as $\cycles$.
We then aim to avoid any of the assignments in the bidding cycle $(j,j',i,i')$.
We introduce slack variables, $\cyclevars_{jj'ii'} \geq 0$, for each bidding cycle, and add a penalty $\cyclepen$ in Equation~\eqref{eq:cyclepenalty} for every assignment that belongs to a bidding cycle.\footnote{In the conference and this paper's experiments, we used an alternate formulation of the constraint which only penalized bidding cycles that led to assignment cycles. In this formulation, slack variables $\cyclevars_{jj'}$ were created for each pair of reviewers $(j,j')$ involved in a bidding cycle, and the constraint was enforced with $\assignmatentry_{ij} + \assignmatentry_{i'j'} \leq (1 + \cyclevars_{jj'})$. We present the above formulation instead because, on reflection, we recommend penalizing both halves of a cycle separately.}
\begin{align}
    \label{eq:cycleconstraints}
    \assignmatentry_{ij} \leq& \  \cyclevars_{jj'ii'} , \forall j,j',i,i' \in \cycles \\
    \label{eq:cyclepenalty}
    \obj^{\text{cy}} =&  \sum\limits_{j,j',i,i' \in \cycles} \cyclepen \cyclevars_{jj'ii'} 
\end{align}



Finally, our overall assignment problem can be stated as:
\begin{align*}
    &\text{max} \ \obj = \obj^\text{match} + \obj^{\text{cap}}+ \obj^{\text{sen}}+ \obj^{\text{co}} + \obj^{\text{reg}} + \obj^{\text{cy}} \\
    &\text{subject to } \constraints.
\end{align*}

\subsection{Solving the Review Assignment problem}
\label{sec: MIP_solving}

Given our formulation, instantiating a variable $\assignmatentry_{ij}$ for each possible (reviewer $j$, paper $i$) combination, along with all of the associated slack variables and constraints, would lead to a MIP too large to store in memory, let alone to solve. We therefore applied column (variable) generation, only creating variables for a subset of all possible assignments and employed row (constraint) generation. 

\subsubsection{Generation of Assignment Variables} We only created $x_{ij}$
variables that were likely to be relevant (thereby restricting the space of possible assignments; implicitly, non-generated variables are set to 0). There was little point in considering variables that correspond to $(i,j)$ pairs with low matching scores, as these would represent poor matches.  In rare cases, this may result in papers receiving fewer than the desired number of reviews. When the MIP is infeasible in this way, one can iterate (perform column generation) by generating more matching variables for the corresponding papers. However, at AAAI we assigned matches to such papers manually, as the algorithm is unlikely to do much better than a random assignment when given only low-scoring alternatives. 
We did not create variables for any reviewer--paper pair $(i,j)$ with $\scorematrixentry_{ij} < 0.15$.\footnote{We note that this suggests a refinement to the bidding procedure: it is only necessary to show reviewers papers for which a sufficiently positive bid could bring their score above the threshold. This could result in large time savings for reviewers in the bidding phase, as it can be hard to sift through an entire conference worth of papers.} 

We created variables as follows: for each paper, we created a variable for the $k$ highest scoring PC, SPC, and AC reviewers. Similarly, for each PC/SPC/AC member, we created a variable (if it did not exist already) for their highest scoring $k/5k/10k$ papers. 
$k$ can be iteratively expanded as time allows (perhaps in a targeted way, such as increasing $k$ for underutilized reviewers) until the dense matrix is recovered. However, our MIP was computationally feasible at high enough $k$ that further increases to $k$ offered only negligible improvements, so we did not iteratively perform column generation. 

\subsubsection{Generation of Coauthor Constraints}\label{subsubsec:rowgen}
Even given our reduced $k$, there were too many coauthor constraints to include (the number scales with the product of the number of $\assignmatentry_{ij}$ variables and the number of reviewers).
Instead, we started solving the problem with no coauthor constraints. After reaching an initial solution, we identified violated coauthor constraints and added them explicitly to the MIP. We repeated this process until we were satisfied with the number of violated constraints; in practice the number of violations plateaued after eight iterations. 
Pseudocode 
is given as Algorithm~\ref{alg:itersolve}.

\begin{algorithm}[H]
\begin{algorithmic}[1]
\small
\caption{Assignment Variable Sparsification and Constraint Generation}
\label{alg:itersolve}
\STATE Let $\coreviewerset_{\text{master}}$ contain all coauthor constraints
\STATE Let $\assignmat_k$ be the set of $x_{ij}$ variables obtained by sampling a constant multiple of the top $k$ papers for every reviewer and top $k$ reviewers for every paper

\STATE $i \gets 0$
\STATE $\coreviewerset_{i} \gets \{\}$ \COMMENT{Store of violated coauthor constraints up to iteration $i$, initially empty}
\WHILE{True}
    \STATE Solve the MIP defined by $\assignmat_{k}$ and $\coreviewerset_i$. Store the solution in $\assignmat^*$
    \STATE Identify all new coauthor violations $\coreviewerset_{\text{new}}$ in $\assignmat^*$ using $\coreviewerset_{\text{master}}$
    \IF{$V_{\text{new}}$ is empty or time has run out}
        \STATE Return $\assignmat^*$
    \ENDIF
    \STATE $\coreviewerset_{i+1} \gets \coreviewerset_{i} \cup \coreviewerset_{\text{new}}$
    \STATE $i \gets i+1$
\ENDWHILE
\end{algorithmic}
\end{algorithm}

Lastly, to reduce computation time, we did not attempt to find an optimal solution to the MIP.
Instead, we set an absolute MIP-gap tolerance of $20$. Since aggscore is on a scale between $0$ and $1$ and we make on the order of tens of thousands of assignments, being suboptimal by at most $20$ objective function units is relatively benign. The result was that we found a very nearly optimal solution much more quickly.


\section{Two-Phase Reviewer Assignment}
\label{sec: two-phase}
The quality of papers submitted to large conferences such as AAAI 2021 varies vastly. An influential experiment by organizers of the 2014 NeurIPS conference \citep{lawrence&al17} split reviewing between two independent Program Committees and had them both review 10\% of submissions. It found that the strongest papers (a small set) and the weakest papers (a much larger set) could be reliably identified, but that many papers close to the decision boundary were accepted by one Program Committee and rejected by the other. In the years since, conference organizers have sought to reallocate reviewing resources from papers that are nearly certain to be rejected to papers that have a realistic chance of acceptance, to improve review quality for the latter group. A popular approach is to employ simple heuristics designed to detect low-quality papers, a process known as ``summary rejection'' or ``desk rejection''.
For example, in IJCAI 2020, Area Chairs were asked to spend a short amount of time skimming each paper to decide whether it deserved a more careful review \citep{bessiere&al20}.
Neurips 2020 employed a similar system: over three weeks, Area Chairs skimmed over 9,000 papers to identify obvious rejects and senior Area Chairs cross-checked these choices; 11\% of submissions received summary rejects \cite{yuan&al20}. Given the size of AI conferences, such summary rejection procedures are time consuming for Area Chairs. Furthermore, they are likely to be noisy and also may reflect unconscious biases against superficial properties of a paper, meaning that they may not be reliable enough to eliminate more than a relatively small fraction of papers. Finally, they tend to be unpopular with authors, who dislike having their paper rejected via an opaque process that produces no reviews. For these reasons, NeurIPS decided not to employ such an approach in 2021 on the basis of negative feedback received in 2020.

We advocate an alternative early-rejection approach that simultaneously concentrates the conference's reviewing budget on papers close to the decision boundary; provides meaningful feedback to authors of early-rejected papers; and early-rejects few papers that could ultimately have been accepted (see our evaluation below). The key idea is to break reviewing into two phases. {In Phase 1} each paper is assigned 2 reviewers. Papers that receive two sufficiently-high-confidence reviews recommending rejection are rejected at this stage, with the authors immediately receiving these full reviews and being offered no opportunity for rebuttal. The process then proceeds to Phase 2, where two or more additional reviewers are assigned to each of the papers that remain. After the second round of reviews, rebuttals are solicited from authors and reviewers from both phases are asked to read rebuttals and each other's reviews, to engage in a discussion mediated by SPCs and ACs, and ultimately to revise their reviews accordingly. Program Chairs then make decisions based on recommendations from ACs.

The key rationale for this system is that, given the low acceptance rates at AI conferences, papers for which the first two reviews are both confident and negative have a very low chance of eventual acceptance. Because reviews are much more careful than quick perusals by ACs, this system is also able to reject a larger fraction of papers than the summary rejection approaches employed by IJCAI, NeurIPS, and other conferences. The resulting reviewing resources can be devoted to papers with a larger chance of acceptance, reducing variance in these recommendations. In the experimental section below, we offer evidence both that papers rejected in Phase 1 were indeed likely to be rejected if given more reviews, and that additional reviews performed in Phase 2 indeed reduced variance in confidence-weighted average scores. 

It is also worth mentioning three additional benefits of the Two-Phase approach. First, inevitably some reviewers fail to complete their assignments, others report low confidence, and still others point out a need for additional reviews by specialists in particular topics. When this occurs in Phase 1, new reviewers can be found without needing to resort to ad hoc, manual approaches that scale badly to large conferences. 
Second, the existence of a second reviewer--paper matching period at the beginning of Phase 2 makes it possible to fast-track high-quality rejected submissions from another conference, using their previous reviews and some numerical constraint on their average scores to take the place of the reviews that would have been obtained in Phase 1. We deployed such a ``fast-track'' in AAAI 2021 for papers that received NeurIPS and EMNLP reviews that scored them roughly in the top half of submissions. Third, many authors appreciate receiving rejection notifications quickly so that they can resubmit their work to another venue without waiting for the whole review process to conclude.

Finally, we note two key drawbacks of Two-Phase reviewing: reviewers have to be chased twice, increasing workload for Program Chairs, ACs, and SPCs; and authors rejected in Phase 1 have no opportunity to rebut reviews that they consider erroneous.

While we designed our Two-Phase reviewing system independently, we have since learned that various versions of the idea were previously implemented at several other CS conferences. To the best of our knowledge, the first conference that split reviewing into two phases was ICML 2009~\cite{Bottou}. According to the program chairs, the motivation was not increasing the number of reviews assigned to strong papers, as it was for us, but rather a concern that papers would be rejected by reviewers who did not fully understand them, due to exponential growth in submissions that had led to a pool of papers larger than the pool of trusted reviewers. ICML09’s system differed from ours in at least two key ways: in Phase 1, authors in ICML 2009 were asked to indicate a preference for an Area Chair who would oversee their paper in Phase 2 and help to choose reviewers; papers were allowed to submit a rebuttal in response to Phase 1 reviews. Two-Phase reviewing was discontinued at ICML in 2011, apparently because the highly manual process became both logistically unwieldy and excessively time consuming as the conference continued to grow.

We are aware of two other (non-AI) conferences that also employed Two-Phase reviewing: ASPLOS, and OOPSLA \cite{Adve2010,10.1145/2641638.2641648}. Notably, in OOPSLA13~\cite{10.1145/2641638.2641648}, authors of accepted Phase 1 papers were asked to revise their papers based on Phase 1 reviews; Phase 2 reviews then assessed the revised versions. 
ASPLOS21~\cite{Kozyrakis2021} required authors to submit a two-page extended abstract in Phase 1; about 55\% of the papers were then promoted to Phase 2 and asked to submit a full paper. In addition to these conferences, \citet{Hans} describe a Two-Phase reviewing process for PLDI conferences, in which Phase 1 decisions are made based on at least three reviews and authors of the papers that are candidates for Phase 1 rejection are invited to submit author responses before a final decision is made for their papers. The author response period for the rest of the papers happens at the end of Phase 2.

AAAI 2021's successful deployment of Two-Phase reviewing has already begun to show some impact of its own. On the practical side, our Two-Phase design was maintained by AAAI 2022 and was newly adopted by both IJCAI 2022 and ICML 2022. The design has also received academic study. A recent paper by \citet{https://doi.org/10.48550/arxiv.2108.06371} investigated how reviewers should be divided across the two phases to maximize the overall matching score in a conference, showing that the problem is NP-Hard and empirically verifying that a simplified \textit{random-split} strategy of randomly reserving reviewers for Phase 2 gives a near-optimal assignment on real conference data.

\section{Evaluations}
\label{sec: evaluations}

\subsection{Analysis of Match Quality}
\label{sec: scoring_function_evals}
We deployed the reviewer--paper matching approach described above for the AAAI 2021 conference. In this section, we analyze a dataset of 6,729 submissions\footnote{For this analysis we removed papers that were withdrawn, desk rejected, or rejected as dual submissions. We also excluded all papers from the AI for Social Impact track as these papers were matched via a somewhat different process. We included papers that were only submitted in the second phase.}, 
the specific reviewer matching we obtained for them,
and 
the resulting paper
reviews
from 8,072 reviewers in the main conference's PC\footnote{In AAAI 2021, Program Chairs, ACs, and SPCs manually assigned about 4.5\% of the final reviewer assignment. In fact, many of these were suggested by area chairs in Phase 2 based on the reviews in Phase 1. In our evaluations, we only considered matches that were automatically assigned to avoid confounding factors of manual matches.}. Unfortunately, we are unable to release the data used for any analyses because data about a conference review process is inherently sensitive, but we summarize our findings in what follows.
We used the following parameters\footnote{The values of these parameters express tradeoffs between violating a constraint and match quality, and hence induce the reviewer--paper matching problem's objective function. There is no ``right'' way to set them: they reflect Program Chairs' subjective answers to questions like the value of having senior reviewers allocated to each paper and the risk of authors forming bidding cycles. This paper does not experimentally investigate these parameter settings, but we note that other conference organizers could conceivably want to set them differently.} (see Section~\ref{subsec: MIP_def} for definitions): $\regionreward = 0.1$, $|\regions|=5$, $\cyclepen = -0.05$, $p^\text{co}(1) = -0.3$, $p^\text{co}(2) = -0.2$, $\seniorityreward = 4$, $\text{TargetSeniority}=4$, $\text{MinSeniority}=0$. In Phase 1, we set reviewer capacities to 3 for PCs, and set $\gamma_\text{pc} =2$, $\gamma_\text{spc} = \gamma_\text{ac} = 1$; we increased PC capacity to 4, and set $\gamma_\text{pc} =4$, $\gamma_\text{spc} = \gamma_\text{ac} = 0$ in Phase 2. We only made assignments to SPCs and ACs in Phase 1 except for the fast track papers for which we set $\gamma_\text{spc} = \gamma_\text{ac} = 1$. We set reviewer capacities to 24 and 60 for SPCs and ACs respectively. We set $W=5$ with $\capacity_{j\cdot}= [8,12,16,20,24] \forall j \in \spcs$ and $\capacity_{j\cdot}= [20,30,40,50,60]$ for $j \in \acs$, with $\capacityslackpenalty_w(\cdot)=-0.05, 1 \le w \le (W-1)$ and a very high penalty of $-0.5$ for $\capacityslackpenalty_W(\cdot)$ for both SPCs and ACs, so that the system never assigned beyond $\capacity_{jW}=\capacity_j$. 


We begin by asking whether we made good matches in the real conference. Before we begin our analysis, we note that evaluating match quality is conceptually tricky. Of course, the gold standard is to ask Area Chairs to manually ensure that matchings make sense; it is also possible to monitor social media for outrage \cite{lawrence_2014}. Our focus here was to go beyond such subjective measures, performing a data-driven and quantitative evaluation along as many dimensions as possible. What follows is, to our knowledge, the most comprehensive post-hoc evaluation of a Computer Science conference's match quality. (Indeed, we note that such analyses are usually limited only to the graphs that appear in Program Chairs' presentations at conference business meetings. Of course, the confidentiality of reviewing data typically prevents others from conducting further, in-depth analyses.)


\begin{figure*}
    \centering
    \includegraphics[width=.6\textwidth]{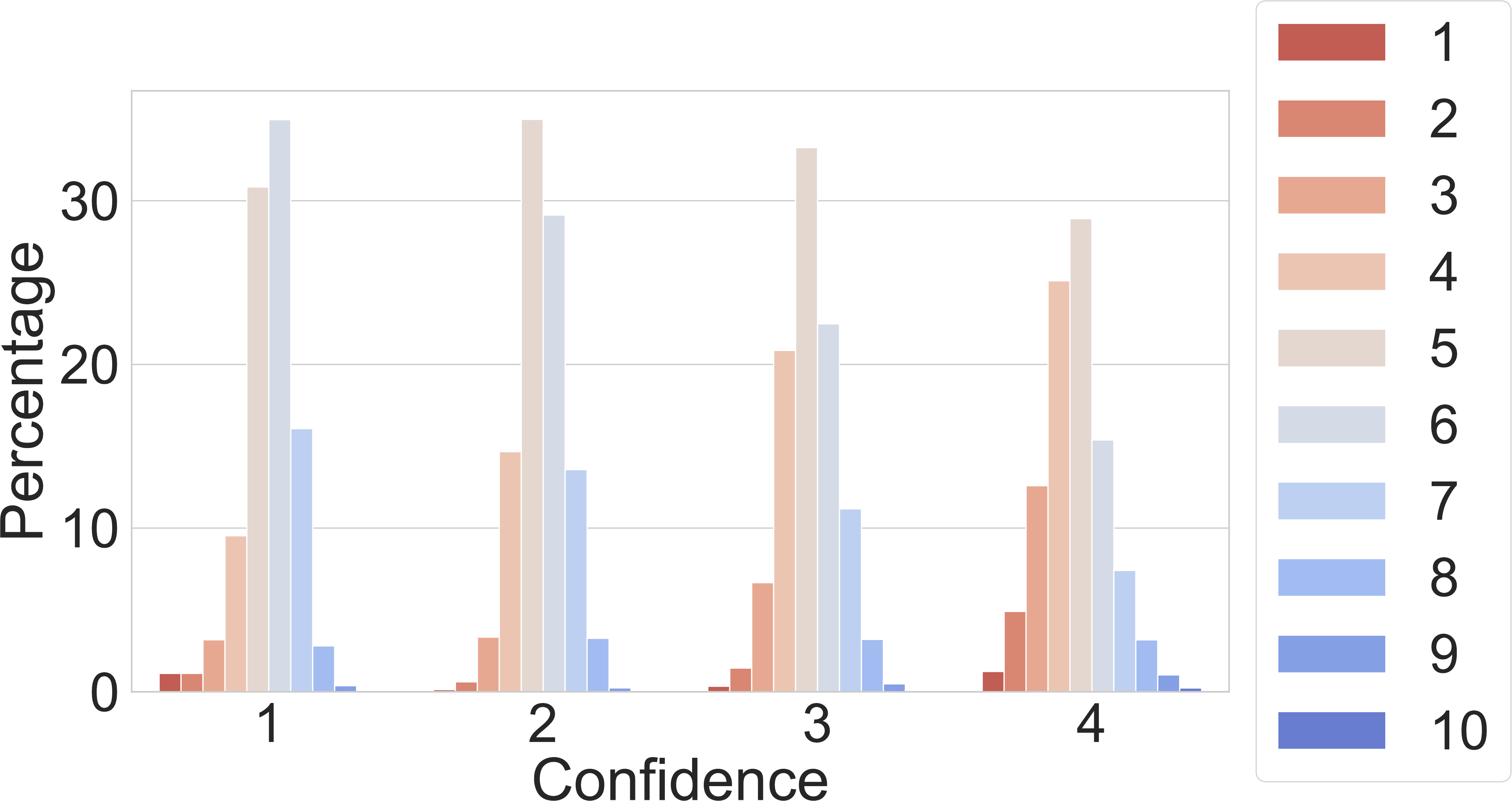}
    \caption{Paper score distribution by confidence of reviewer. More confident reviewers gave fewer 5s and 6s.\label{fig:confidence_v_score}}
\end{figure*}
\begin{figure*}
    \centering
  \includegraphics[width=.6\textwidth]{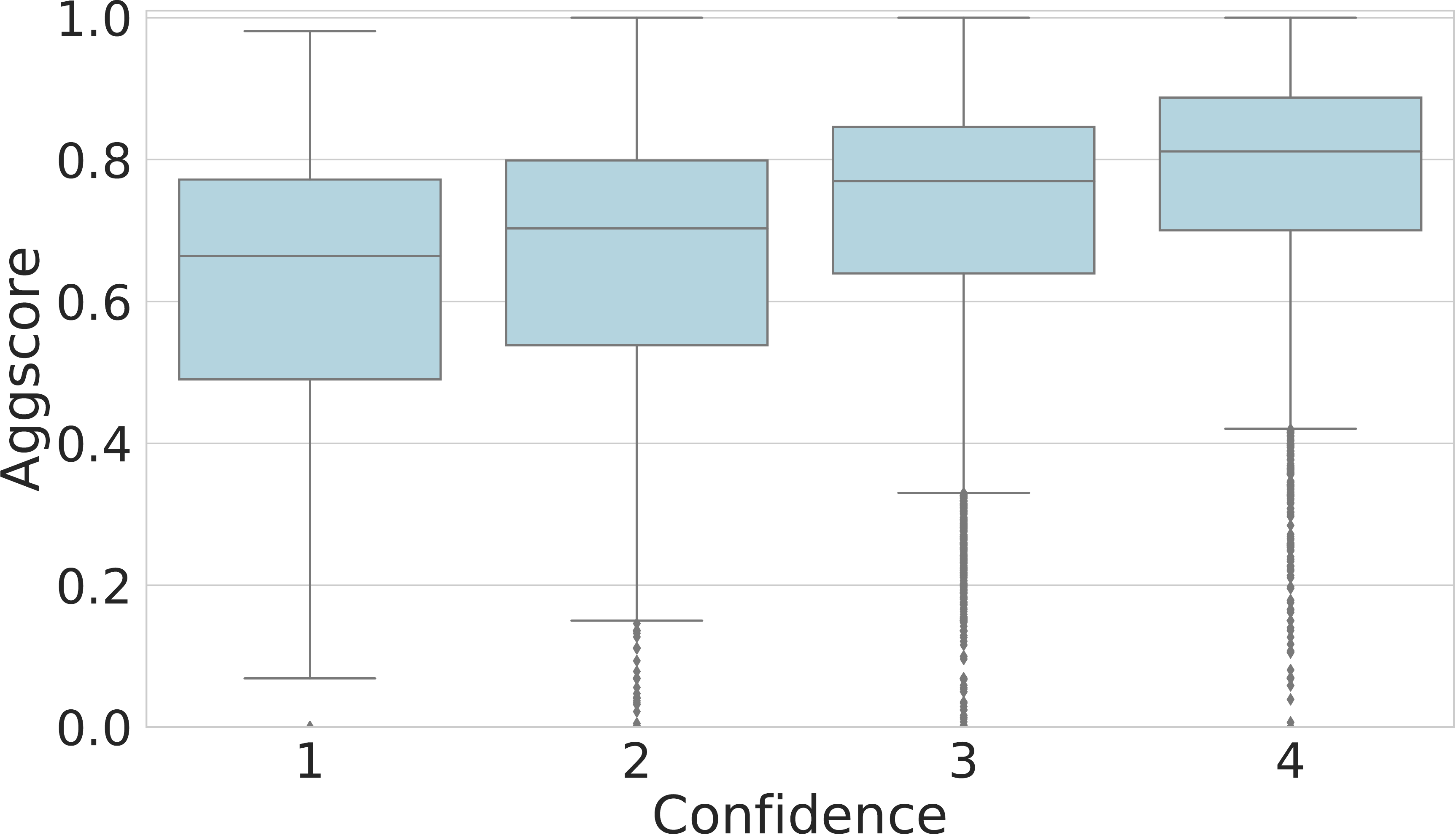}
    \caption{Reported confidence of an eventual review vs. predicted aggscore for all reviewer--paper pairs matched at AAAI 2021.\label{fig:confidence_v_aggscore}}
\end{figure*}


\subsubsection{Did our scoring function capture reviewer expertise?} First, we investigated whether reviewers actually bid on papers for which we assigned them high ``expertise'' scores, leveraging the fact that reviewers tend to bid positively on papers that they are capable of reviewing. 
Considering the set of reviewers who submitted at least one bid, we looked at all matched reviewer--paper pairs that we assigned base scores (i.e., aggscores before taking bids into account) greater than $0.75$. We found that 29\% of these pairs were accompanied by \textit{eager} bids, 38\% \textit{willing}, and 3\% \textit{in a pinch} (the remainder no bids). 
We see this as evidence that high base scores indeed correlated with high match quality. However, we acknowledge that the bidding interface is a possible confounding factor here---e.g., users could view papers in descending order of  their TPMS scores (one of the inputs to aggscore).

Second, we observe that the distribution of overall scores given by reviewers to papers (which ranged between 1 and 10) noticeably changed as a function of self-reported confidence.\footnote{Confidence levels could be reported as ``Reviewer made an educated guess'', ``Reviewer is broadly knowledgeable, but not an expert in the area'', ``Reviewer is knowledgeable in the area of the paper'', or ``Reviewer is an expert in the specific topic of the paper''.} Figure \ref{fig:confidence_v_score} shows that more confident reviewers tended to give more informative paper scores (fewer borderline 5 and 6 scores). Given that higher confidence reviews were more informative, we examined the relationship between our scoring function and reported reviewer confidence. Figure~\ref{fig:confidence_v_aggscore} shows that our scoring function was positively correlated with confidence: as confidence increased, so did the 25th, 50th, and 75th percentile of scores.
\subsubsection{Assuming that our scoring function predicted review quality, how close did we come on average to assigning the best possible reviewer to each paper?}
We considered the ordinal ranking of reviewers assigned to each paper in terms of aggscore, where rank 1 corresponds to the reviewer with the highest aggscore for that paper. We then looked at the mean reviewer ranking for each paper with respect to the ranks of all of the paper's reviewers.
The first / second / third quartiles of mean reviewer rankings across all papers were 7 / 12 / 28. This shows that papers were often assigned very highly qualified reviewers; note that these averages are taken across all of a paper's reviewers (median 4) and that a single reviewer with a poor rank is able to cause the average ranking to grow large.

\subsubsection{Was our aggscore computation stable under missing data?} 

Recall that aggscore, our paper-reviewer matching score, is a combination of three constituent scores: TPMS, ACL, and SAM (see Section~\ref{subsec:scoreagg}).
At AAAI 2021 these constituent scores were available for $60.25\%$, $96.64\%,$ and $100\%$ of papers respectively. Here we evaluate the stability of our aggscores in cases where ACL and TPMS scores were missing. 
We began with all ``qualified'' (paper, reviewer) pairs ($\scorematrixentry_{ij} \geq 0.15$) for which all of the three scores were available. 
We then computed two proxy aggregate scores, representing the scores we would have computed if either of the TPMS or ACL scores had been missing.

The standard deviation of the changes in aggscores caused by missing data were 0.049 and 0.048
respectively, whereas the mean standard deviation of the complete-data score for a given paper was 0.127.  
Since the variance of the differences between the hidden data aggscores
and the full-data aggscores was much smaller than the average variance in aggscores across reviewers for a paper, the aggscore was still likely to have made a similar assessment of reviewer--paper pairings even in the presence of missing data.




\subsubsection{How many conflicts were identified by our COI detection?}
Let us denote self-reported conflict domains, explicitly stated conflicts, and  conflicts between coauthors of papers submitted to AAAI 2021 as `trivial conflicts'.
Out of the total $2,674,372$ conflicts we found at AAAI 2021, $96.4\%$ were trivial. Of the remaining $3.6\%$ conflicts, $2.8\%$ were due to unreported coauthorship relationships we verified via DBLP. The remaining $0.8\%$ of conflicts were between predicted student--supervisor pairs or predicted students of the same supervisor. Overall, our system added at least one nontrivial conflict to a large majority ($78.8\%$) of submissions.

\subsection{Analysis of MIP Solving}
\label{sec: MIP_scalability_evals}
We now turn to an evaluation of our matching algorithm (described in \Cref{sec: MIP}), including a comparison to the matching algorithm used at AAAI 2020, the previous year's conference. 
For our experiments, we created a full conference dataset containing all submitted AAAI 2021 papers and reviewers analyzed above. We augmented the dataset to include all additional reviewers who were not ultimately assigned any reviews at AAAI 2021, bringing the total number of reviewers up to 8,964. Except where otherwise noted, we evaluated our algorithm in terms of the single-phase matching problem, as this allows us to focus on matching quality and facilitates comparison with prior work. In the same spirit, we required each paper to be assigned four reviews ($\gamma_\text{pc}=4$) and set a limit of five reviews per reviewer ($\capacity_j = 5$). We ran row generation (as described in \Cref{subsubsec:rowgen}) for 10 iterations. We ran all of our experiments on a 16-core machine with Intel Silver 4216 Cascade Lake 2.1GHz processors and 96 GB RAM. For MIP solving, we used CPLEX version 12.9.0.0 with default parameters. 

We note a subtlety that arises when comparing different assignments. Recall that we used an inequality instead of a hard constraint on the number of matches each paper receives, allowing the MIP to assign fewer than $\gamma_\text{pc}$ reviewers to a paper. This can occur for papers for which there are no qualified reviewers or only very poor matches conditional on the rest of the assignment. A consequence is that we can find ourselves needing to make comparisons across assignments in which different numbers of reviewer--paper matches exist. In these cases, we padded all matchings with reviewer--paper matches that contributed nothing to the objective function, as though we had matched the paper to a reviewer having an aggscore of 0 who imposed neither rewards or penalties via soft constraints. Such zero-contribution matches accounted for at most 1\% of the total reviews in any of our matchings reported below. While in reality we addressed all such cases at AAAI 2021 manually, we note that many AAAI 2021 reviewers were ultimately assigned zero papers, meaning that a large pool of available and unused reviewers did exist.

\subsubsection{How did our algorithm compare to the algorithm used in the previous iteration of the conference?}

We compared our matching algorithm to the method used in the conference's previous iteration (AAAI 2020), a flow-based matching algorithm. Rather than simple capacity constraints, each paper and reviewer had a number of weighted \emph{slots} that linearly scaled the contribution of each assignment to prioritize each paper getting some good reviewers and each reviewer getting some good papers. Based on advice from the previous Program Chairs, we set the paper slot weights to be 8/4/1/1 and reviewer slot weights to be 8/2/1/1/1. Their reviewer--paper scoring function differed from ours in two main ways: (1) it did not use ACL scores, and (2) bids scaled scores multiplicatively rather than exponentially. Beyond the scoring function, their matching algorithm did not attempt to satisfy any soft constraints.

Nevertheless, we investigated how effective last year's approach was at limiting soft constraint violations despite not optimizing for them (i.e., we tested how frequently these violations occurred organically at the optimum of a simpler matching problem). We ran both algorithms and evaluated the number of soft constraint violations compared to our approach. For reference, we also performed a similar experiment that compared our AAAI 2021 algorithm to a version of the same algorithm that omitted all soft constraints. We summarize the results in \Cref{tab:constraint_improvement_over_2020}, reporting the percent reduction in the number of:
(a) papers without a senior reviewer (Constraint \ref{cons:seniority});
(b) coauthor violations (Constraint \ref{cons:coauthor});
(c) papers with reviewers from the same region (Constraint \ref{cons:geo}); and
(d) cycle violations (Constraint \ref{cons:cycle}).
We observed that AAAI 2020's approach led to substantially more violations of three of the four soft constraints. Most notably, our algorithm assigned to the same paper 76.8\% fewer pairs (118 compared to 509) of reviewers who were previous coauthors . Conversely, our approach led to a 90\% \emph{increase} in the number of violations of our no-2-cycles constraint, prohibiting pairs of reviewers who bid positively on each other's submissions from being assigned to each other's papers. However, such matchings were quite rare overall (22 for AAAI 2020 and 42 for our approach out of 26,916 reviewer--paper matchings in total).
We believe that the AAAI 2020 method led to fewer 2-cycles because their scoring function weighs positive bids less heavily than ours does and therefore is less likely to match reviewers to papers where a reviewer bid positively.
When using our scoring function, we were able to reduce 2-cycle violations by 84\% using our soft constraint over the baseline version with no soft constraints.

Our algorithm had approximately the same memory footprint as the flow-based approach (96 GB vs 95 GB) despite the additional soft constraints that our method considered: the flow-based method creates an edge for every reviewer--paper pair whereas we only created variables for a subset of qualified reviewers. In terms of runtime, running our algorithm with 10 iterations of row generation required 52.0 wall-clock hours compared to 0.8 needed for the AAAI 2020 approach. Note that our approach is anytime: it can terminate after any number of iterations of row generation. One iteration of our algorithm took 0.9 wall-clock hours and five iterations took 7.3 wall-clock hours.
We were willing to devote a few days to computation, so we find these numbers reasonable. Nevertheless, in what follows we identify some ways to reduce runtimes without sacrificing much in performance (e.g., $5\times$ speedup via decomposing the matching into two stages; $1.7\times$ speedup via reducing $k$ to further sparsify the constraint matrix).
\begin{table*}
\begin{center}
\small
\begin{tabular}{ l| r r}
\toprule
 \textbf{Soft constraint} & \textbf{Baseline (no soft constraints)} & \textbf{AAAI 2020} \\
 \midrule
  \textbf{Constraint \ref{cons:seniority}} (Seniority) & 56.1 & 9.5\\
  \textbf{Constraint \ref{cons:coauthor}} (Coauthorship Distance) & 88.1 & 76.8\\
  \textbf{Constraint \ref{cons:geo}} (Geographic Diversity) & 24.1 & 23.3\\
  \textbf{Constraint \ref{cons:cycle}} (No 2-Cycles) & 84.6 & -90.9\\
  \bottomrule
\end{tabular}
\caption{Percent reduction of soft constraint violations using our approach relative to (1) our approach without soft constraints and (2) last year's matching algorithm for a simulation of a single-phase version of AAAI 2021.}
\label{tab:constraint_improvement_over_2020}
\end{center}
\end{table*}
\subsubsection{What did introducing soft constraints cost us in matching aggscore?}\label{subsec:softeval}
Table \ref{tab:improvement_over_2020} shows that for the most part, with our specific settings of the soft constraint parameters, we were able to have our cake and eat it too. With all of our soft constraints turned on, we only lost 2.04\% mean aggscore. With each soft constraint turned on individually, the Coauthorship Distance and Geographic Diversity constraints reduced mean aggscore the most (1.17\% and 0.85\% respectively) while the Seniority and No 2-Cycles constraints reduced mean aggscore the least (0.05\% and 0.01\% respectively). 
The sum of the aggscore reductions across matchings where each soft constraint was turned on individually was 2.08\% (sum of first four rows in Table \ref{tab:improvement_over_2020}), very similar to the 2.04\% reduction in mean aggscore when all constraints were turned on simultaneously. This suggests that there was very little complementarity across different constraints (e.g., when we reduced Coauthorship Distance violations, we rarely also increased Geographic Diversity as a side effect).

\begin{table*}
\begin{center}
\small
\begin{tabular}{ l r}
\toprule
 \textbf{Soft constraint} & \textbf{\% Aggscore reduction} \\
 \midrule
 \textbf{Constraint \ref{cons:seniority}} (Seniority) & 0.05\\
 \textbf{Constraint \ref{cons:coauthor}} (Coauthorship Distance) & 1.17 \\
 \textbf{Constraint \ref{cons:geo}} (Geographic Diversity) & 0.85\\
 \textbf{Constraint \ref{cons:cycle}} (No 2-Cycles) & 0.01\\
 \midrule
 All constraints & 2.04 \\
  \bottomrule
\end{tabular}
\caption{Percent reduction in aggscore when each soft constraint was turned on individually relative to a base MIP having no soft constraints. The final row displays the reduction when all soft constraints were present.}
\label{tab:improvement_over_2020}
\end{center}
\end{table*}

\subsubsection{How well did the row generation of coauthorship distance constraints perform?}
\label{subsec:roweval}
Ideally we would evaluate the performance of the row generation of coauthorship distance constraints by comparing performance after each number of iterations to the performance of the optimal approach that held all rows of the constraint matrix in memory. Unfortunately, we were unable to compute the optimal objective value (OPT) for our dataset: this is why we needed row generation in the first place! Nevertheless, we can compare to an upper bound on OPT. 
Since each additional coauthorship constraint can only penalize the objective function, the optimal value of the objective function (without coauthorship penalty from ungenerated constraints) can only weakly decrease as these soft constraints are added. Therefore, the objective value we observe after the last iteration of row generation is an upper bound on OPT.
After 10 iterations, we closed the gap between the initial solution before row generation (obtained with $\coreviewerset_0$ in \Cref{alg:itersolve}) and the upper bound on OPT by 77.5\%.


\subsubsection{How did the choice of our column generation parameter $k$ impact the objective?}
\begin{figure*}[t]
\centering
  \includegraphics[width=0.45\textwidth]{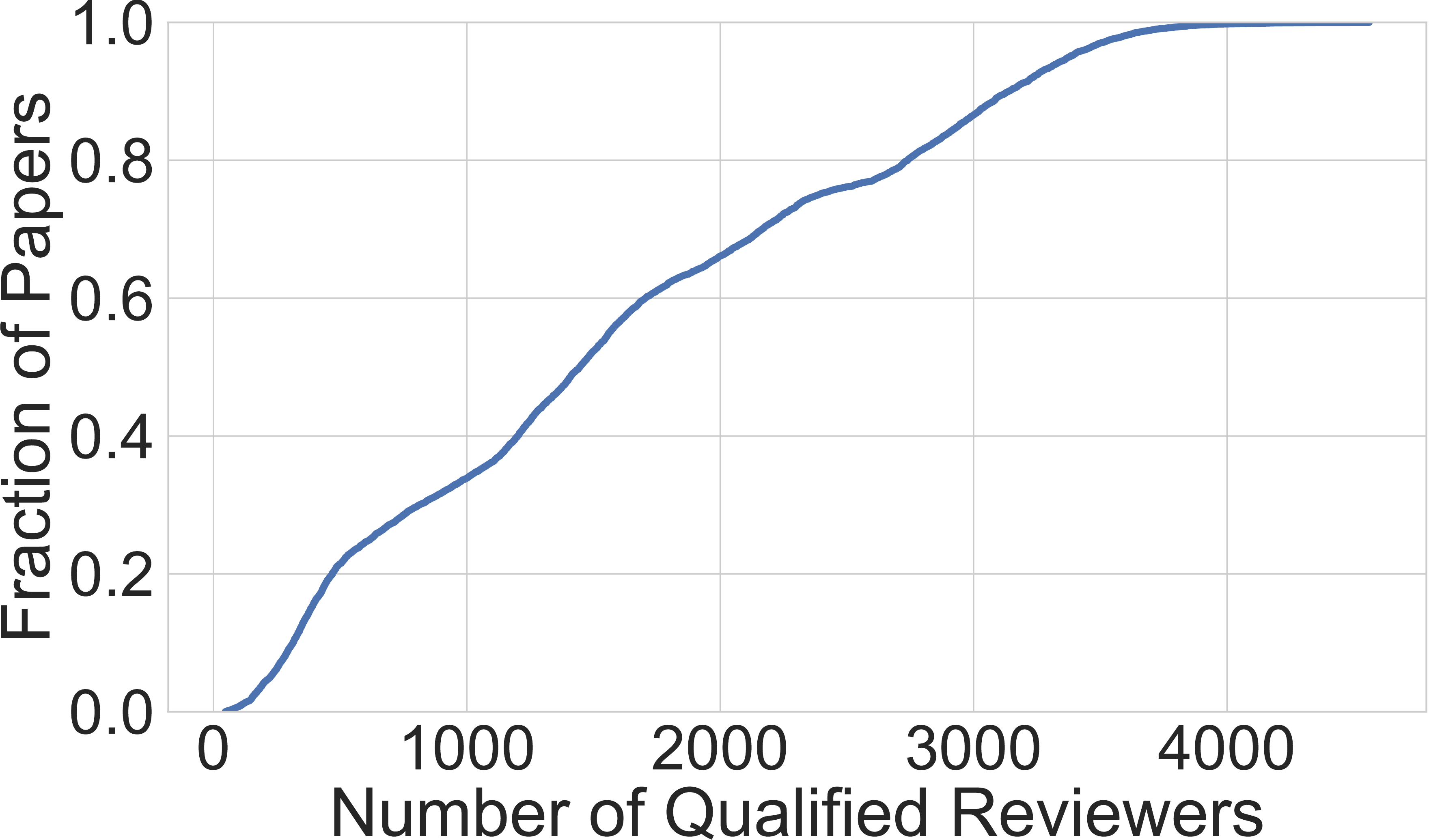}\hspace{.09\textwidth}
  \includegraphics[width=0.45\textwidth]{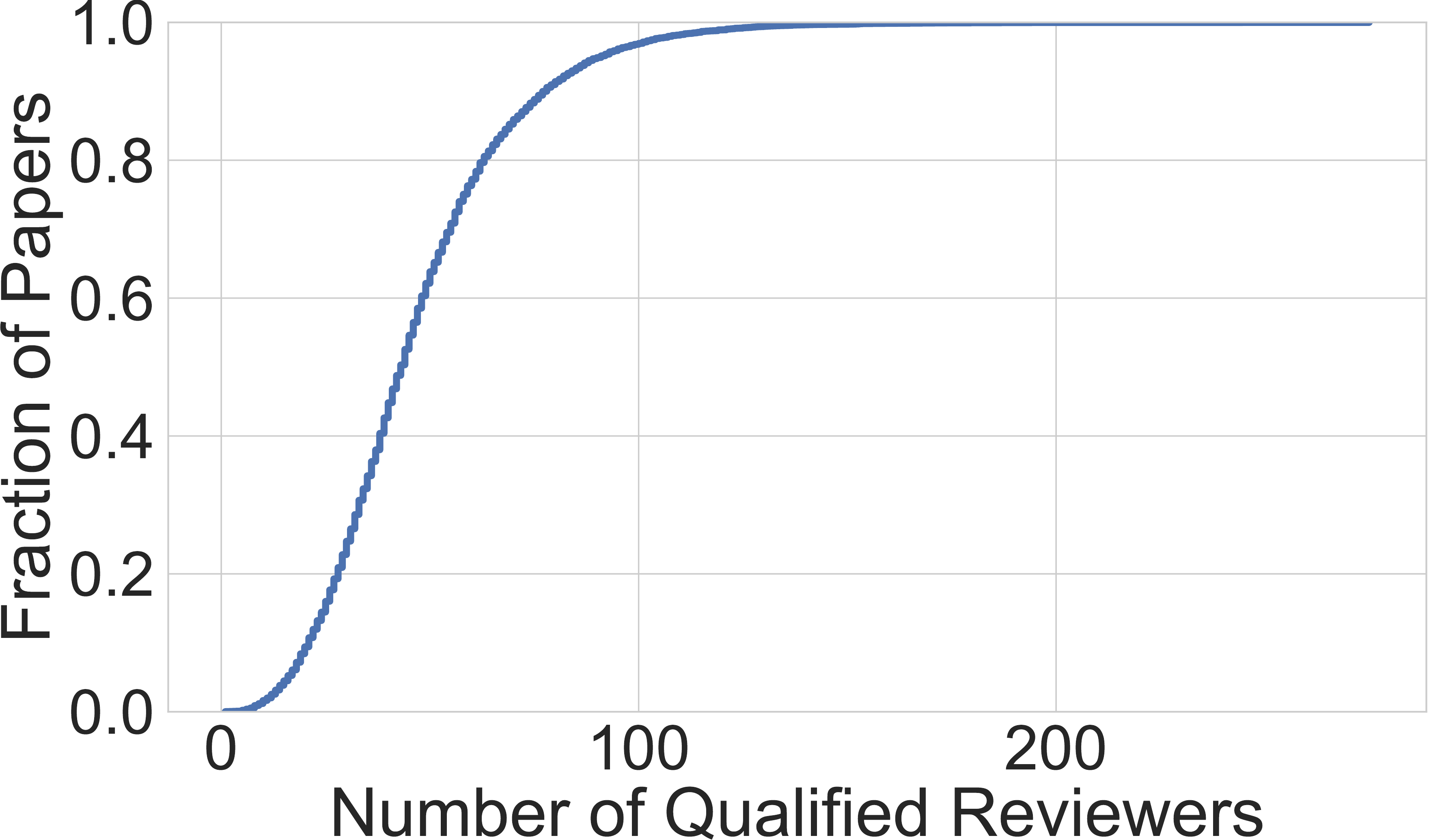}
  \caption{ECDF of the number of ``qualified'' reviewers for each paper (i.e., reviewers with $\scorematrixentry_{ij} \geq 0.15$), considering both all reviewers (Left) and only reviewers who also made a positive bid for the paper (Right). The median paper had 1446 qualified reviewers and 43 qualified reviewers who also placed a positive bid.\label{fig:qualified}}
\end{figure*}

Recall that to reduce the size of our MIP, we created variables only for the most-promising paper--reviewer pairs: for each reviewer, we sampled the $k$ best-scoring papers, and for every paper, we sampled the $k$ best-scoring reviewers. We then filtered out any paper--reviewer pairs scoring below a threshold, keeping only qualified reviewers. We set $k=50$ when creating the match for AAAI 2021. 
\Cref{fig:qualified} shows that after this filtering, the median paper had $\approx$1,500 qualified reviewers. 
Expanding to include variables for all possible qualified assignments (unbounded $k$) would have required much more memory than was available via our computing resources.

We investigated the impact of $k$ on the quality of the resulting matching. Running our algorithm with a much smaller $k$ value ($k=10$), we observed only a 1.8\% loss in the objective but a 1.7 times speedup in performing 10 iterations of row generation, as compared to $k=50$. Increasing $k$ to $100$ had a negligible impact on the objective (0.04\%). We consider it unlikely that significant additional gains would be achieved by even further increases to $k$. 


\subsubsection{What were the tradeoffs between objective value and computational savings with multiple phases?}

In a two-phase conference, about half of the total reviews are assigned in each phase, so the Phase 1 and Phase 2 MIPs have half as many free variables as the MIP for a corresponding single-phase conference.  We should expect solving the two smaller problems to be much faster than solving the larger one, since the running time of MIP solvers tends to scale superlinearly with problem size. However, of course, the solution obtained by solving two phases optimally in succession will yield weakly lower objective value than the optimal solution to the single-phase problem. For example, imagine that each stage assigns two reviews per paper and that for some paper, there are only four qualified reviewers. The two-phase algorithm will not consider that it may need to save two qualified reviewers for the second phase, and may assign these highly specialized reviewers elsewhere, consuming their capacity. We thus empirically evaluated loss in objective value due to running a two-phase version of our conference. In these experiments, we assigned two reviewers per paper in Phase 1, fixed those assignments, and then assigned two more reviewers per paper in Phase 2. We found that the resulting matching had objective value only 0.02\% lower, as compared to the single-phase conference. On the other hand, decomposing the problem in this way led to a $5\times$ speedup.

It occurred to us that this approach of incrementally building up a MIP solution reviewer by reviewer did not require that the conference actually did perform two-phase reviewing. For example, a single-phase conference could use this same two-phase strategy simply to speedup MIP solving, which could make a significant computational difference for conferences significantly larger than AAAI 2021. To investigate this idea, we evaluated a four-phase approach that assigned one reviewer per paper, fixed those assignments, and repeated four times until each paper had four reviewers. As before, we observed very encouraging results: the objective value decreased by only 0.33\% relative to a single-phase conference, whereas MIP solving was accelerated by 28$\times$ (see \Cref{tab:multi-phase})!

\begin{table}
\begin{center}
\small
\begin{tabular}{ l r r}
\toprule
 \textbf{Number of phases} & \textbf{Objective reduction} & \textbf{Speed up}\\
 \midrule
2 & 0.02\%  & 5.4$\times$ \\
4 & 0.33\%  & 28.1$\times$ \\
  \bottomrule
\end{tabular}
\caption{Sacrifice in objective function vs. computational savings of multi-phase conferences relative to single-phase conference.}
\label{tab:multi-phase}
\end{center}
\end{table}









\subsubsection{How did our algorithm scale?}

To understand how our algorithm's performance depended on the size of the conference, we generated 77 synthetic conferences of different sizes by subsampling papers and reviewers (without replacement) from our dataset. In an attempt to maximize the realism of the structure of our simulated conferences, we constructed them based on related keywords. At AAAI 2021, keywords were divided into top-level categories (e.g., Machine Learning) and bottom-level categories (e.g., Learning Theory); papers and reviewers could select one top-level and multiple bottom-level keywords.

Our generator works as follows. We begin with an initial bottom-level keyword and then initialize our conference to contain all papers associated with that keyword and a set of reviewers associated with the corresponding top-level keyword proportional to the fraction of papers belonging to the bottom-level keyword within that top-level area. We store a set of keywords associated with our conference as the generator runs. In each iteration, we add to our set of keywords the keyword belonging to most papers in our current set of papers that is not yet part of our tracked set of keywords. Papers corresponding to this keyword and additional reviewers are then added accordingly. 
Every time the number of papers grew by a constant factor, we created a new synthetic conference containing the current set of papers and reviewers and added it to our set of samples.
For diversity, we ran our generator on three trajectories, each seeded with a different keyword: one from game theory, one from computer vision, and one from constraint satisfaction, because each of these areas involves a substantially  distinct set of reviewers and authors and each has its own dedicated conferences. We also included the full AAAI 2021  dataset as our final datapoint, giving us a total of 78 simulated conferences.

\begin{figure*}[t]
\centering
    \includegraphics[width=.6\textwidth]{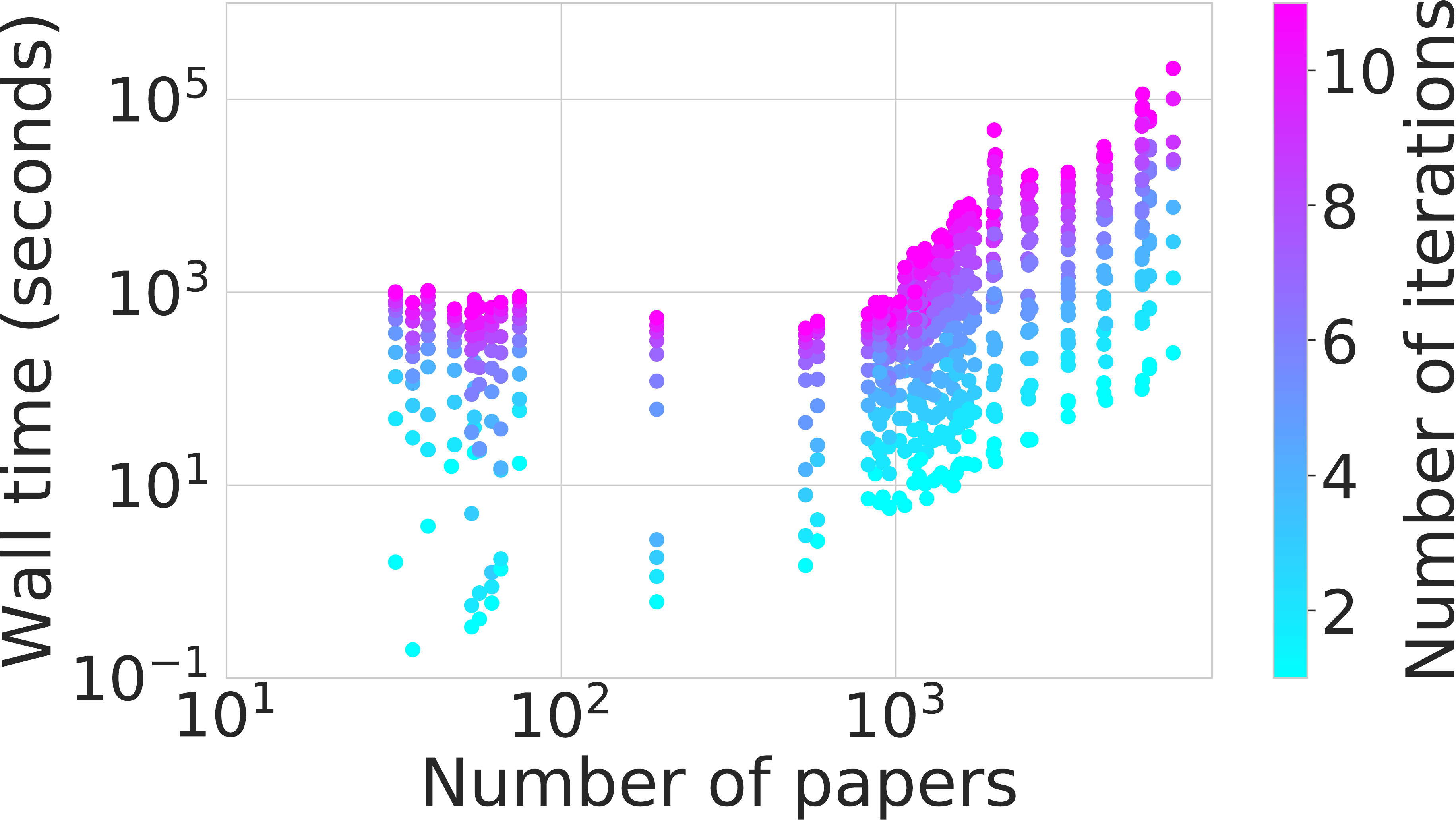}
    \caption{Running time vs. conference size (number of papers); points represent (conference, iteration) pairs.\label{fig:scale}}
\end{figure*}

We then turned to studying how our row generation approach scaled. Figure~\ref{fig:scale} shows cumulative wall-clock time for each (conference, iteration) point. Points are coloured by iteration. For conferences with fewer than 1,000 papers, we could always complete 10 iterations in less than an hour. Interestingly, the smallest conferences were harder to optimize than medium-sized conferences. We speculate that this occurred because some soft constraints (e.g., coauthor / cycle constraints) were harder to satisfy when the set of papers and reviewers were more densely connected. Running time appeared to scale exponentially as the number of papers grew beyond 1,000, with the biggest size taking a few hours per iteration.

\begin{figure*}[t]
\centering
  \includegraphics[width=.6\textwidth]{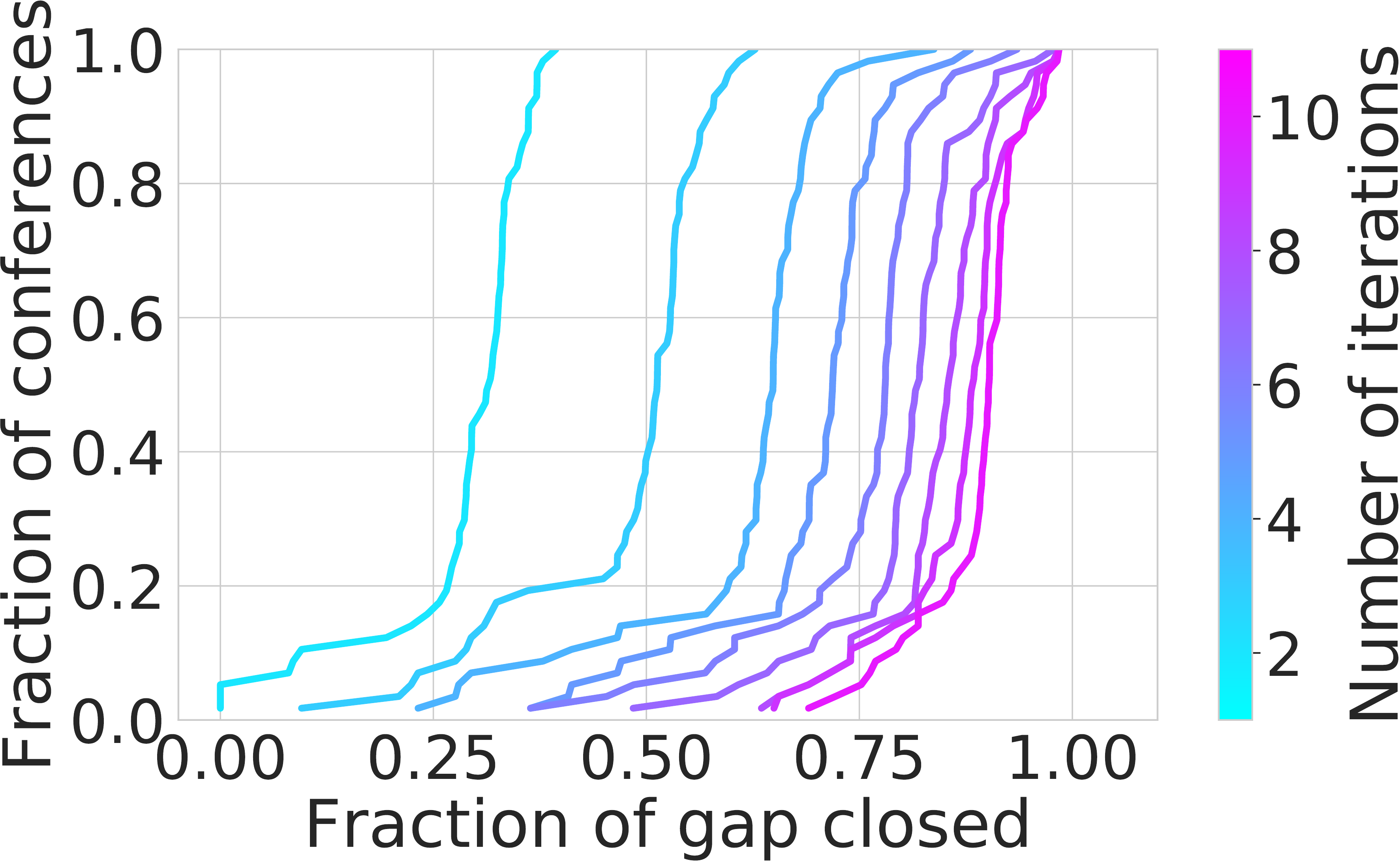}
  \caption{ECDF of \% of gap closed (relative to our upper bound) between the first and $k^{th}$ iteration of row generation.\label{fig:optgap}}
\end{figure*}

\subsubsection{How did our results on optimality and aggscore reduction translate to our generated conferences?}

\Cref{fig:optgap} shows ECDFs of how much of the gap we closed between the initial solution before row generation and the upper bound on OPT described earlier in \Cref{subsec:roweval}. After just one iteration of row generation, we closed at least 25\% of the gap for 80\% of the generated conferences.  After 10 iterations, we closed 75\% of the gap for over 90\% of conferences. The trend in which these ECDFs become closer and closer together after successive iterations led us to believe that we were approaching optimality, and that much of the remaining 
gap after 10 iterations was due to looseness in our upper bound on OPT.
\begin{figure*}[t]
\centering
  \includegraphics[width=.6\textwidth]{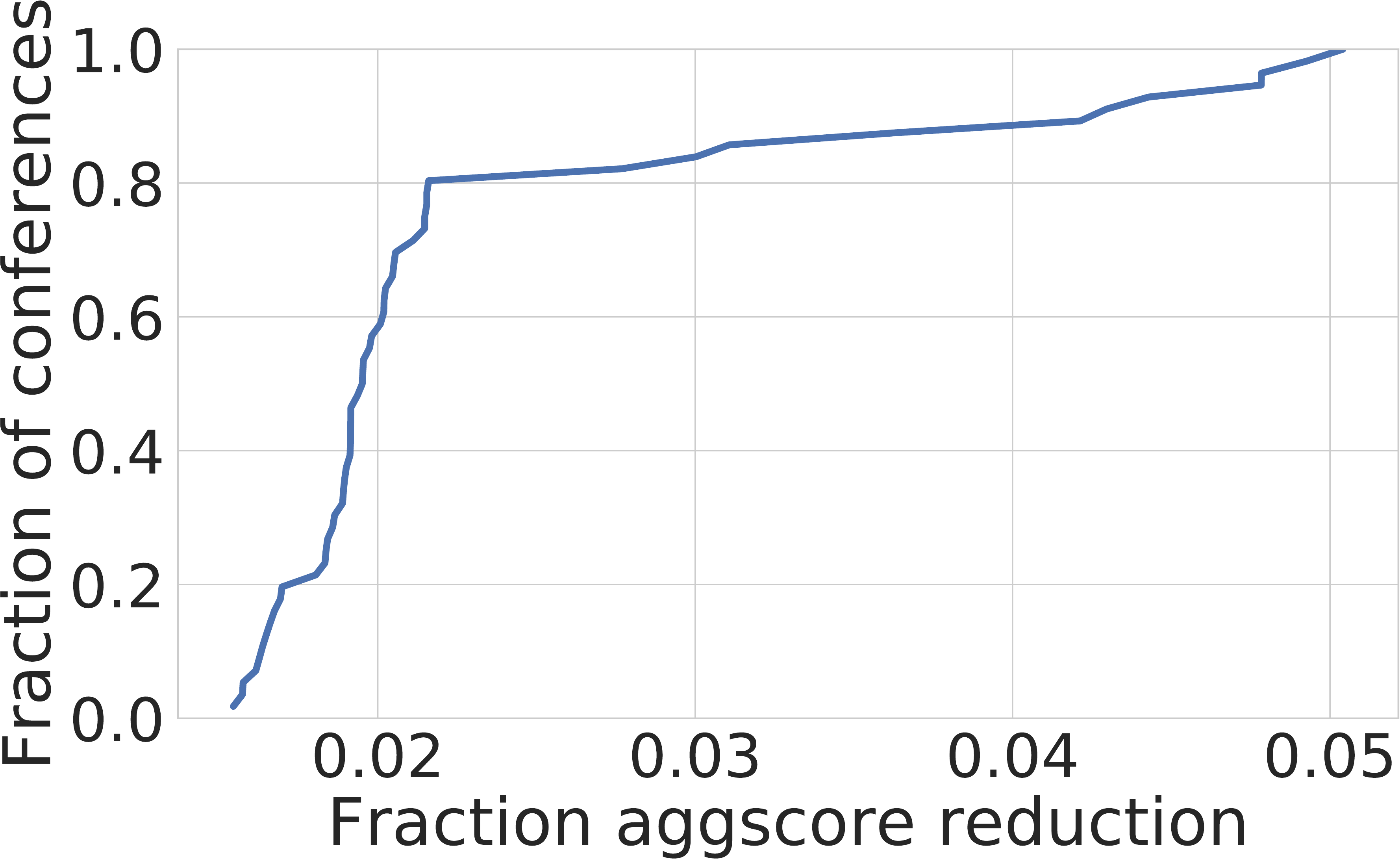}
  \caption{ECDF of the fractional reduction in mean aggscore upon adding soft constraints.}\label{fig:agg_score_reduction}
\end{figure*}

Similar to the results previously reported about our full conference in \Cref{subsec:softeval}, we observed that introducing soft constraints led to only small reductions in aggscore across our 77 variable-sized conferences. For 60\% of these conferences, we lost less than 2\% in aggscore; at worst, we lost 5\%.  
\Cref{fig:agg_score_reduction} shows an ECDF of the reduction in aggscore across all the generated conferences.

\subsection{Evaluation of Two-Phase Reviewing}
\label{sec: two_phase_evals}

\subsubsection{How prevalent were false negatives?}\label{sec:false-negatives} 
The risk we were most concerned about arising from our Two-Phase design was the possibility that papers rejected in Phase 1 might have been accepted if they had received additional reviews, an opportunity for rebuttal, and/or discussion; we call such papers ``false negatives''. The cleanest way to estimate the probability with which we falsely rejected such papers in Phase 1 would have been for us to conduct a randomized experiment: randomly promoting some fraction of papers that would otherwise have been rejected in Phase 1 and observing the result. We did not perform this experiment, but the real data provides a very similar natural experiment. Specifically, we promoted papers to Phase 2 without considering their scores in two cases: (1) when one or more reviews were missing; and (2) when their reviewers had low confidence. In total, 231 papers were promoted in this manner, 16 of which were eventually accepted. Each of these papers eventually received four or more high-confidence reviews, allowing us to calculate the probability that two randomly selected reviews would have met the criteria for Phase 1 rejection. This allowed us to estimate what fraction of Phase 1 rejections might eventually have been accepted. In our data this probability was  $2.9\%$, suggesting that Phase 1 rejections included very few false negatives.


\subsubsection{Did Phase 1 reviewers participate in discussion after Phase 2?} There was a relatively long delay between the end of Phase 1 and the start of the discussion period (in our case, almost a month), leading to some public speculation that Phase 1 reviewers might have tended to be less active in the discussion phase~\cite[e.g.,][]{Kambhampati}. We found little evidence of such a trend: Phase 2 papers received a discussion post from 55\% of Phase 1 reviewers and likewise 55\% of Phase 2 reviewers. We did find that reviewers who reviewed only in Phase 1 (who were responsible for a relatively small fraction of reviews overall) had a slightly smaller (52\%) rate of participation in the discussion. We further note that Two-Phase reviewing is not the only possible explanation for this small discrepancy: e.g., this group may also have contained a smaller fraction of highly qualified reviewers.





\subsubsection{How many additional reviews were gained?} AAAI 2021 received 7,133 full paper submissions to the first phase. 
We were able to reject 2,615 (about $37\%$) of the submitted papers in Phase 1, leaving us with a surplus of {2,615} reviews (relative to AAAI's previous practice of assigning 3 reviews to each paper) to spread amongst the remaining papers. 
 This contributed towards assigning
 at least four reviewers to every main track paper in Phase 2 and at least 3 reviews to the 737 fast track submissions, out of which 721 submissions received more than 4 reviews.
    
\subsubsection{How important was it to have additional reviews?} 
\begin{figure*}
    \centering
  \includegraphics[width=.6\textwidth]{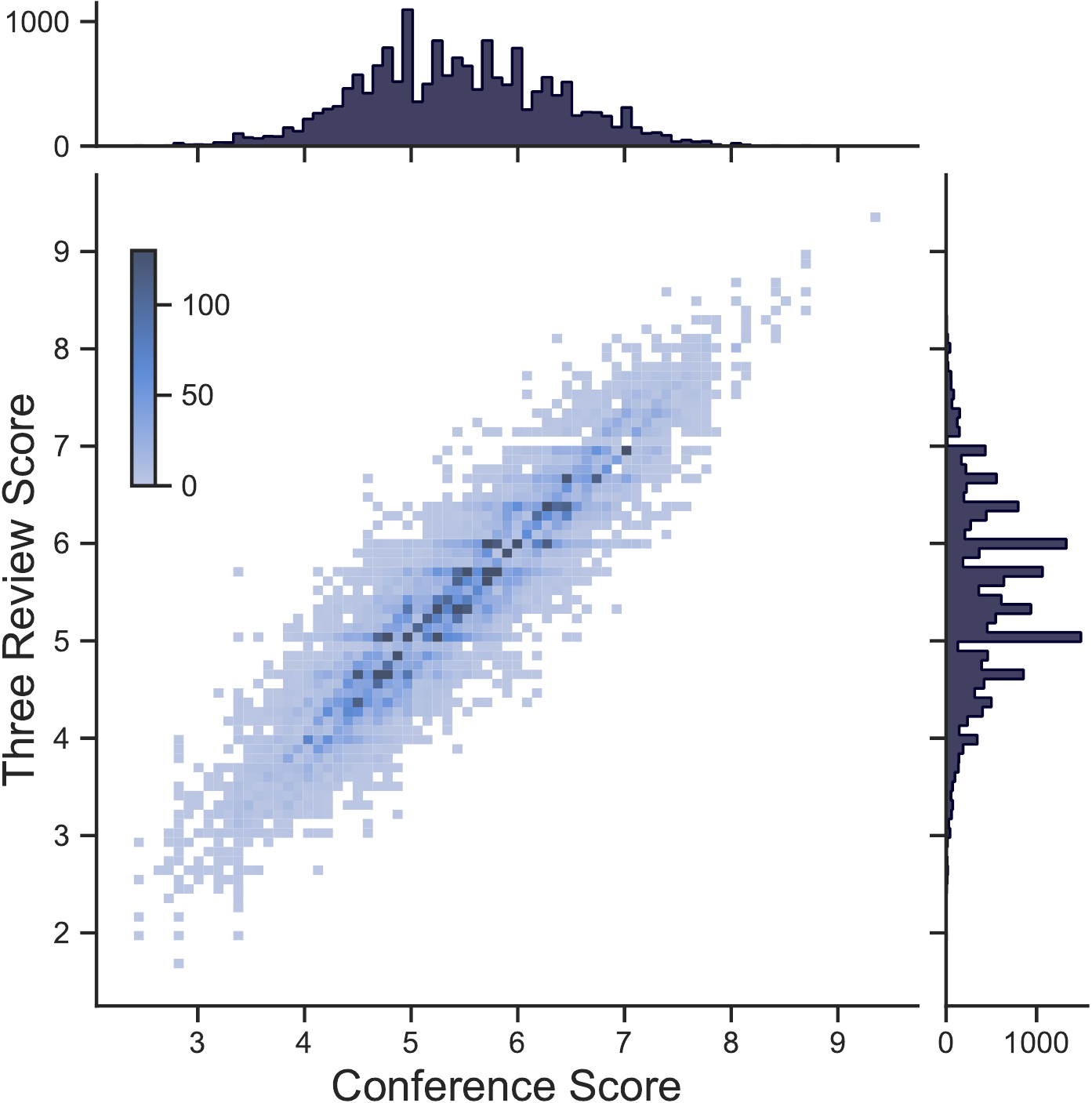}
  \caption{Confidence-weighted average scores for Phase 2 papers based on all reviews vs.\ all combinations of 3 reviews. Marginal densities are plotted on the sidelines. \label{fig:aaai_score_variance}}
\end{figure*}

For every Phase 2 paper, we sampled every subset of 3 reviews that it might have received if a subset of the same reviewers had been assigned in a single-phase conference and calculated the confidence-weighted average score that the paper would have received in this scenario. (Confidence ranged between 1 (low) and 4 (high); corresponding weights were 0.25, 0.5, 0.75, and 1.) 
We plot the result in Figure~\ref{fig:aaai_score_variance}. Each point corresponds to a 3-review scenario for a single paper. The $x$ axis gives the paper's score in the AAAI 2021 conference (where it received more than 3 reviews); the $y$ axis gives the spectrum of 3-review scores. The decision boundary for acceptance fell around 6.4, though many papers below this threshold were accepted and many above this threshold were rejected. Nevertheless, variance in the score is a good proxy for variance in reviewer support for the paper. Figure~\ref{fig:aaai_score_variance} shows that for papers that are close to the decision boundary, the weighted average score could vary by up to 2 points around the threshold depending on which 3-reviewers were picked, implying that decisions for these papers could be very uncertain with only 3 reviewers. Therefore, we conclude that the additional reviews enabled by two-phase reviewing helped Program Chairs to make more confident decisions, specially about papers that are close to the decision boundary. 


\subsubsection{Did the adoption of Two-Phase reviewing result in a higher bar for acceptance by ACs/SPCs?}
One concern that has been publicly raised about two-phase reviewing is that it might depress acceptance rates because ACs and SPCs may aim to accept the same fraction of Phase 2 papers as in previous years, not considering the large fraction of papers already rejected in Phase 1  \cite[e.g.,][]{Walsh2022}. Our analysis of AC/SPC recommendations shows that this was not the case in AAAI 2021. ACs submitted 3,825 ratings on a scale of \textit{Reject, Lean Reject, Lean Accept, Weak Accept, Accept, and Best Paper}. Considering the last four scores as recommendations for acceptance, we found that ACs recommended accepting $37.6\%$ of Phase 2 papers, corresponding to an overall acceptance rate of $22.2\%$. Similarly, SPCs submitted 3,873 ratings on a scale from 1 to 9.\footnote{The text prompts given for each score were as follows: (1) Trivial, wrong, or known; (2) Strong reject: will fight to get it rejected; (3) Clear reject; (4) Reject; (5) Below threshold of acceptance; (6) Above threshold of acceptance; (7) Accept; (8) Clear Accept (Top 50\% accepted papers (est.)); and (9) Accept: will fight to get it accepted.} Considering scores above 5 as accept recommendations, SPCs recommended accepting $37.5\%$ of Phase 2 papers, corresponding to an overall acceptance rate of $22.3\%$.
Both of these percentages were higher than the conference's actual acceptance rate of $21\%$ and almost the same as the average acceptance rate of $22.9\%$ over the last 5 years~\cite{openresearch}. We therefore see no evidence to support the claim that introducing two-phase reviewing leads to lower overall acceptance rates.

\subsubsection{Did reviewers behave differently in Phase 2?}
\begin{figure*}
    \centering
    \includegraphics[width=.6\textwidth]{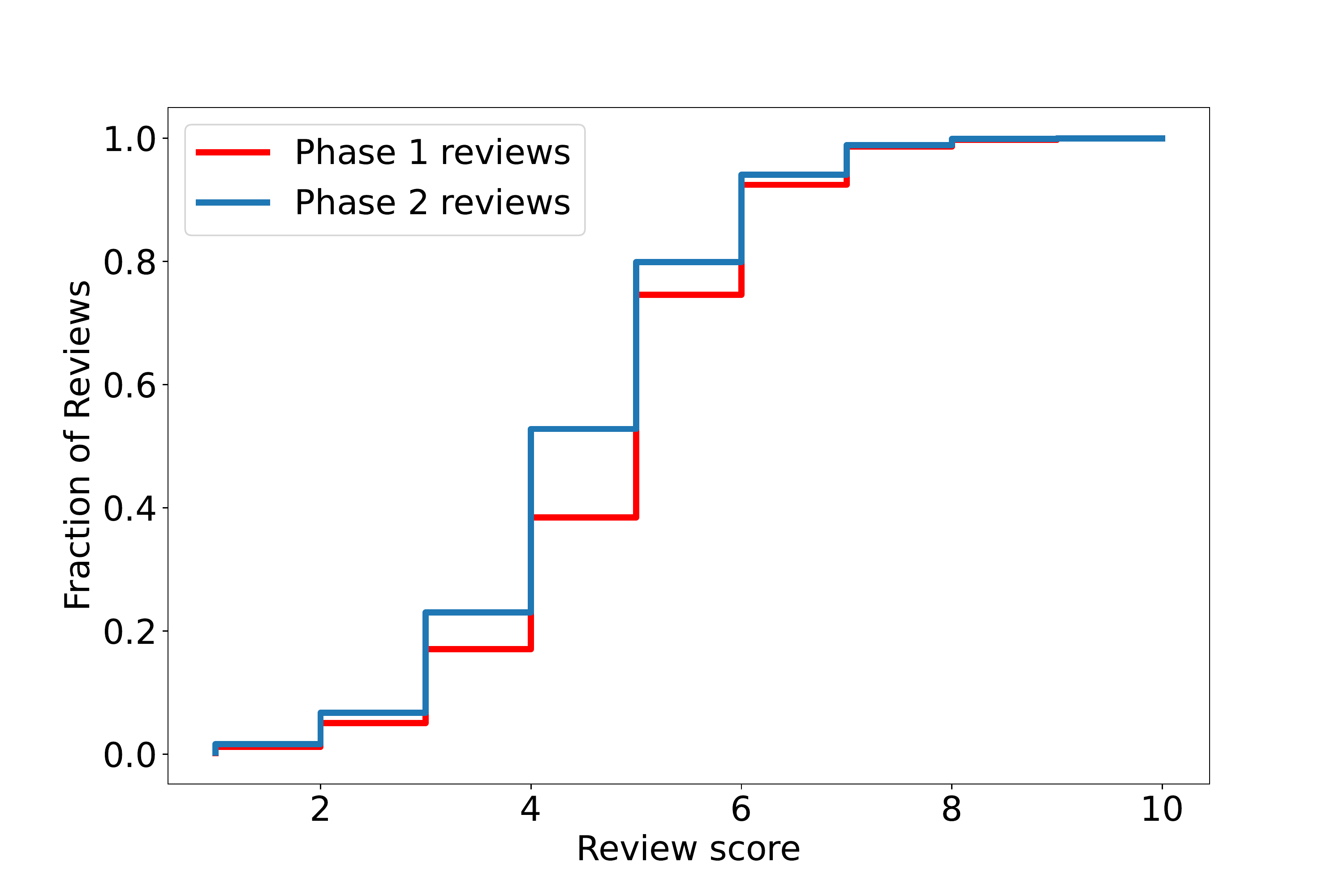}
    \caption{ECDF of the review scores by review phase. Phase 2 reviews tend to be more negative than Phase 1 reviews.\label{fig:ecdf}}
\end{figure*}

In AAAI 2021, Phase 2 reviews tended to be (slightly) more negative than Phase 1 reviews, with Phase 1 pre-rebuttal scores averaging 0.29 higher than Phase 2 pre-rebuttal scores on the 10-point scale (see Figure~\ref{fig:ecdf} for an ECDF).  We used the Wilcoxon signed-rank test \cite{Rey&neuh11} to check whether pairs of Phase 1 and Phase 2 reviews given to the same paper came  from the same distribution (null hypothesis) or whether Phase 2 reviews tended to be more negative (alternative hypothesis). We ran the test with a significance level of 0.01 on a dataset whose entries were every possible combination of a Phase 1 review score and a Phase 2 review score for the same paper. (Thus, we only considered papers which were reviewed in Phase 2.)
We used pre-rebuttal scores 
to make sure that reviewers' scores were not influenced by reading author responses and other reviews, after which review scores tended to become more correlated). 
Over 6160 pairs of scores in our dataset
, the test had a $p$-value of $7.57 \times e^{-69}$, causing us to reject the null hypothesis that reviewing was identical across phases. 

There are various reasons why reviewers might have been more negative in Phase 2: they may have applied a higher standard, knowing this was the final reviewing round; they may have changed their perceptions about quality thresholds after their experiences reviewing in Phase 1; more qualified reviewers might have been identified in Phase 2 and these reviewers may have had more critical views. 

However, a fourth explanation occurred to us, which the data can tell us something about: reviewers may have behaved exactly as they did in Phase 1, but the change in scores might have arisen due to selection bias arising from the Two-Phase structure. 
To see how this might have happened, imagine that each reviewer's score is an unbiased sample from a Gaussian centered on the ``true score'' of the paper. 
While reviewing in this model is unbiased on expectation, performing reviewing in two phases adds bias. That is, if both reviewers in Phase 1 draw samples above the paper's true score, the paper will get more reviews in Phase 2, which on expectation will be more negative. Conversely, if both reviewers in Phase 1 draw samples well below the paper's true score, we'd again expect additional reviews to revert to the mean, but we will not see those additional reviews because the paper will be rejected in Phase 1. This asymmetry means that we should expect reviews in the second phase to be more negative on average for purely statistical reasons; observing such bias is not necessarily cause for alarm.

To investigate the impact of this selection bias effect, we considered a refined dataset of papers that had at most 1 negative review (review score of 5 or below), and hence would not have been rejected in Phase 1 regardless of the order in which their reviewers were chosen; there turned out to be 1583 papers. We again ran the Wilcoxon signed-rank test comparing Phase 1 scores to Phase 2 scores and obtained a $p$-value of $0.57$: we did not have enough evidence to reject the null hypothesis that the distributions of Phase 1 scores and Phase 2 scores were the same. This indicates that these papers were unlikely to have been promoted from Phase 1 purely due to noise.

We now turn to a more quantitative analysis. To see how much of the 0.29-point average score gap between reviews in the two phases could plausibly be explained by selection bias, we needed to estimate the variability of reviewers' process of scoring papers. We thus constructed a simple statistical model of our setting. We assumed that each paper $p$ has a true score $s_p \in \mathbb{R} $ which is drawn from a Gaussian distribution $\mathcal{N}(\mu_s, \sigma^2_s)$ where $\mu_s$ and $\sigma^2_s$ indicate our prior belief on the average and variance of true scores. 
When asked to review paper $p$, reviewer $r$ draws an unbiased sample $o^p_r$ from a normal distribution $\mathcal{N}(s_{p}, \sigma^2)$ around $s_p$, the true score of paper $p$, and with the variance of $\sigma^2$. We note that when we use thousands of observations to fit a handful of parameters, priors are overwhelmed and hyperparameter choices become nearly irrelevant. Nevertheless, performing Bayesian inference requires the selection of priors. We therefore estimated $\mu_s=5$ and $\sigma^2_s=1$ as the empirical mean and variance of overall paper scores and assumed that the prior over $\frac{1}{\sigma}$ was $\GammaDist(\alpha_\tau, \beta_\tau)$, setting $\alpha_\tau= \beta_\tau = 1$.

We used Gibbs sampling to estimate $\sigma$ for reviewers and $s_p$ for every $p$. After taking the average over the Gibbs samples, our inference estimated $\sigma$ to be $1.3$.
We then ran a simulation in which we used our graphical model to generate 6,723 true paper scores\footnote{6,723 is the total number of papers in our database that were either rejected or accepted based on their reviews in Phase 1 or Phase 2---i.e., excluding papers that left the system for other reasons, such as being withdrawn by their authors or being desk rejected because of double submission, violation of author anonymity, exceeding page limits, etc.} from $\mathcal{N}(5, 1)$ and review scores that followed the noise model defined by $o_{p}^r \sim \mathcal{N}(s_{p}, 1.3^2)$. 
We then simulated Phase 2 by filtering out any paper that received two reject recommendations (received two review scores below 4.5) from its first two sampled review scores (in Phase 1). Considering the remaining (Phase 2) papers in our synthetic experiment, we observed that Phase 1 reviews were $0.18$ higher than Phase 2 reviews on average. Overall, this (simplistic) analysis explains about two thirds of the gap that we saw in our real data. We conclude that selection bias likely made reviewing in the two phases appear more different than it really was, but that it is furthermore plausible that reviewers did behave at least somewhat differently in the two phases. 



\section{Conclusion}
\label{sec: conclusion}
This paper has presented a novel method for reviewer--paper matching that is scalable to large conferences and more robust to malicious behavior. Our formulation is based on a mixed-integer program that combines a scoring function (which itself combines three match scores and reviewer bids) with various soft constraints encouraging the optimizer to pick, for each paper, a set of reviewers that is geographically diverse, includes at least one senior reviewer, and does not include co-authors. Several preprocessing steps predict new conflicts of interest alongside known ones, and reviewer bids are also processed to undermine malicious bids. A Two-Phase reviewing scheme uses available reviewing resources more efficiently by allocating fewer reviews to papers that have low acceptance probability. We performed extensive evaluation and post-hoc analysis to demonstrate the value of our methods. We have publicly released our reviewer--paper matching software for further use by other conferences, and indeed this software is already in use at ICML 2022. Furthermore, the two-phase reviewing methodology and various other novel elements of the design described here have been adopted by AAAI 2022 and IJCAI 2022.

\bibliographystyle{unsrtnat} 
\bibliography{references}

\end{document}